\documentclass[journal]{IEEEtran}
\usepackage{xcolor,soul,framed} 

\colorlet{shadecolor}{yellow}
\usepackage[pdftex]{graphicx}
\graphicspath{{../pdf/}{../jpeg/}}
\DeclareGraphicsExtensions{.pdf,.jpeg,.png}

\usepackage[cmex10]{amsmath}
\usepackage{cite}
\usepackage{rotating}
\usepackage{hyperref} 
\usepackage{tablefootnote}
\usepackage{array}
\usepackage{mdwmath}
\usepackage{amsmath}
\usepackage{amsfonts}
\usepackage{amssymb}
\usepackage{mdwtab}
\usepackage{eqparbox}
\usepackage{url}
\usepackage{xcolor}
\usepackage{comment}
\usepackage{footnote}
\usepackage{multirow}
\usepackage{booktabs}
\usepackage{graphicx}
\usepackage{caption}
\usepackage{algorithm, algorithmicx}
\usepackage{algpseudocode}

\usepackage{float}
\usepackage{placeins}

\usepackage{url} 
\usepackage{array, makecell}

\usepackage{tikz}

\usetikzlibrary{fit,positioning}
\usetikzlibrary{bayesnet}
\usetikzlibrary{arrows}

\usepackage{flushend}

\begin{document}
\bstctlcite{IEEEexample:BSTcontrol}
    \title{Onboard Satellite Image Classification for Earth Observation: A Comparative Study of ViT Models}
    \author{\IEEEauthorblockN{Thanh-Dung Le,~\IEEEmembership{Senior Member,~IEEE}, Vu Nguyen Ha,~\IEEEmembership{Senior Member,~IEEE}, Ti Ti Nguyen,~\IEEEmembership{Member,~IEEE}, Duc-Dung Tran,~\IEEEmembership{Member,~IEEE}, Hung Nguyen-Kha, Luis M. Garces-Socarras,~\IEEEmembership{Member,~IEEE}, 
    Juan Carlos Merlano-Duncan,~\IEEEmembership{Senior Member,~IEEE}, 
    Symeon Chatzinotas,~\IEEEmembership{Fellow,~IEEE} 
    \vspace{-3mm}
    }


    \thanks{This work was funded by the Luxembourg National Research Fund (FNR) under the SENTRY project (grant reference C23/IS/18073708/SENTRY).}

    \thanks{Thanh-Dung Le, Vu Nguyen Ha, Ti Ti Nguyen, Duc-Dung Tran, Hung Nguyen-Kha, Luis M. Garces-Socarras, Juan Carlos Merlano-Duncan, Symeon Chatzinotas are with the Interdisciplinary Centre for Security, Reliability, and Trust (SnT), University of Luxembourg, Luxembourg (Corresponding author. Email: thanh-dung.le@tamucc.edu).} 
}

\markboth{Submitted to IEEE Transactions on Geoscience and Remote Sensing, 2026.
}{Thanh-Dung Le \MakeLowercase{\textit{et al.}}: }

 \maketitle

\begin{abstract}
Remote sensing (RS) image classification (IC) is central to Earth observation (EO), and Transformer-based architectures with large-scale pre-training now rival the convolutional neural networks (CNNs) that long dominated it. Deploying such models onboard satellites, however, imposes constraints that standard benchmarks ignore: strict power and memory budgets, and inference on imagery degraded by sensor noise, platform motion, and lossy downlink transmission. This study identifies the most effective pre-trained model for onboard land-use classification by evaluating accuracy, computational efficiency, and robustness jointly rather than in isolation. Following a train-on-ground, infer-onboard workflow, we compare 14 backbones, covering CNN, ResNet, and compact Transformer baselines trained from scratch as well as pre-trained Vision Transformers (ViTs), on the EuroSAT and PatternNet benchmarks, and stress them with Gaussian noise and motion blur at five severity levels as well as an end-to-end DVB-S2(X) transmission chain combining channel impairments with JPEG compression. Pre-trained ViTs consistently outperform models trained from scratch at lower computational cost and with greater noise resilience. MobileViTV2 attains the highest clean-data accuracy (99.09\%), whereas EfficientViT-M2 offers the best overall trade-off: 98.76\% accuracy, precision, and recall on EuroSAT (99.52\% on PatternNet), a compact footprint (203.53~MFLOPs, 38.19~MB), fast training (1{,}000\,s) and inference (10\,s), the strongest corruption robustness (overall score 0.79) with the most graceful degradation under transmission loss, and markedly lower power, 63.35\% below MobileViTV2 (79.23\,W) and 73.33\% below SwinTransformer (108.90\,W). EfficientViT-M2 is therefore the most suitable choice for reliable, energy-efficient onboard RS-IC. Reproducible code for data augmentation, noisy-data generation, training, and inference is openly available from our shared GitHub repository \footnote{
\scriptsize
\url{https://github.com/ltdung/SnT-SENTRY}}.

\end{abstract}


\IEEEpeerreviewmaketitle

\vspace{-2mm}

\section{Introduction}

\IEEEPARstart{T}{he} increasing demand for satellites to support EO and RS missions has driven a rapid expansion in the number of Low Earth Orbit (LEO) satellites \cite{sadek2021new}. These missions are critical for various applications, including environmental monitoring, disaster response, precision agriculture, and scientific research, where high-frequency, high-resolution data is essential. A key component of EO systems is RS-IC, which plays a fundamental role in analyzing satellite-acquired data. Traditionally, IC in RS has relied on CNNs and other DL techniques, which have been widely validated and proven effective for processing and classifying RS data with high accuracy and efficiency \cite{boualleg2019remote, xu2021lightweight, li2020classification}.

However, recent advances have shifted toward Transformer-based architectures, which surpass conventional CNNs in several IC tasks, including RS applications \cite{xu2022vision, li2024vision}. Large-scale pre-trained models, first in natural language processing with BERT \cite{DevlinCLT19} and later in vision with ViTs \cite{khan2022transformers, liu2023survey}, learn rich representations that fine-tune to new tasks with minimal training, driven by greater compute, data, and architectural efficiency \cite{han2021pre}. Yet their high power and resource demands constrain deployment on onboard satellite systems (OSS) for real-time decision-making. This is compounded by a key operational bottleneck for LEO satellites: dependence on ground stations for downlink, whose short visibility windows cause prolonged connectivity loss and delay timely access to mission-critical observation data for end users \cite{al2022survey}.

Satellite Internet Providers such as SpaceX and OneWeb can mitigate this with near-continuous connectivity and on-demand access \cite{chougrani2024connecting}, but connectivity alone cannot meet the low-latency, real-time demands of modern EO. This motivates onboard processing that autonomously analyzes data, prioritizes critical information, and acts immediately, for example by refocusing on the next revisit, without delayed intervention from ground-segment operators \cite{fontanesi2023artificial}.

Consequently, most onboard neural networks (NNs) are deliberately lightweight. The $\Phi$-Sat-1 mission ran the CNN-based CloudScout for onboard cloud detection on an Intel Movidius Myriad 2 VPU, the first DL deployed on a satellite \cite{giuffrida2021varphi}, and $\Phi$-Sat-2 used a convolutional autoencoder for image compression \cite{guerrisi2023artificial}. These cases illustrate how restricted power and processing budgets limit the deployment of more complex Transformer architectures onboard. Yet CNN-based models, while efficient, often miss the intricate patterns and contextual dependencies needed for more advanced RS tasks.

\begin{figure*}[!ht]
    \centering
    \includegraphics[width=1\textwidth]{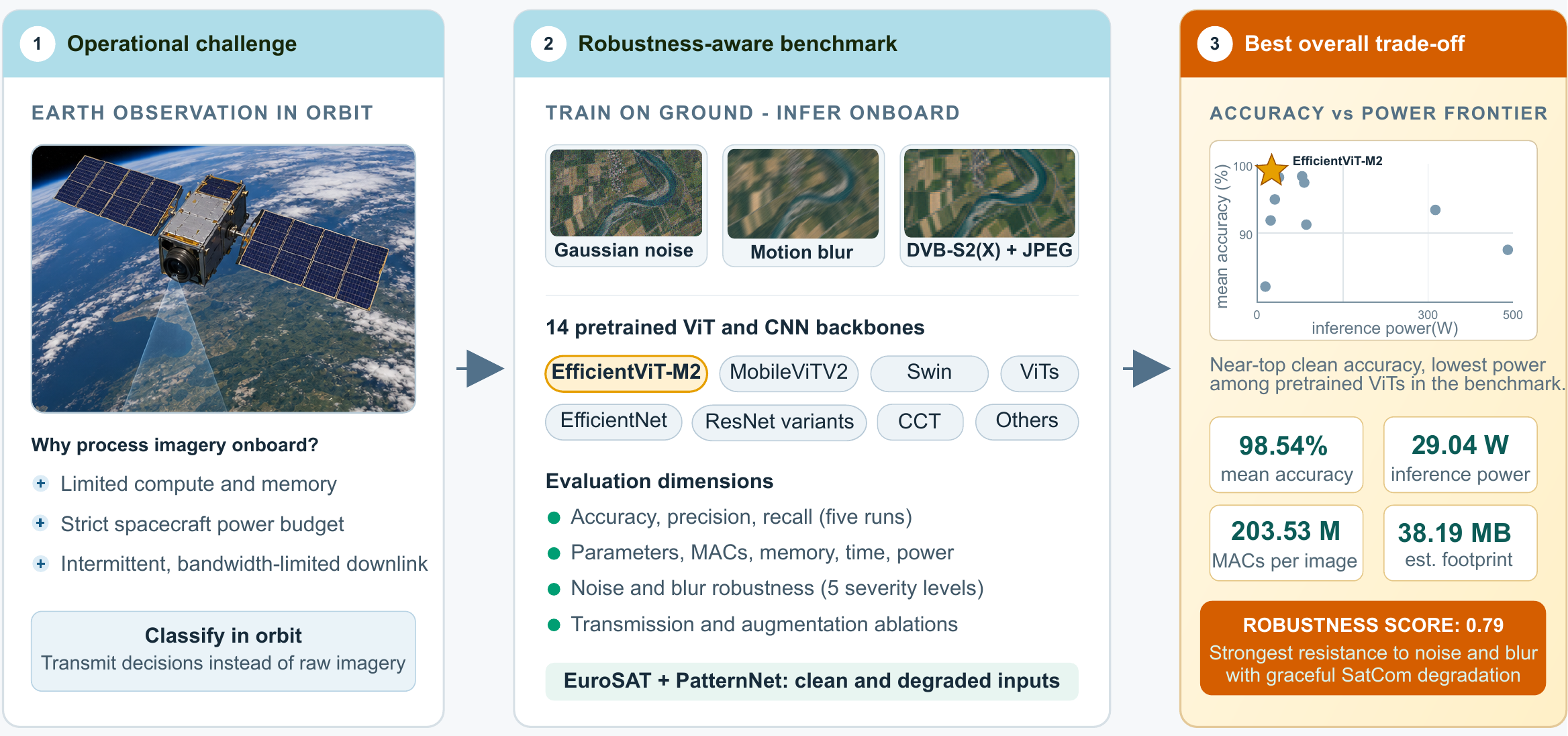}
    \captionsetup{font=small}
    \caption{Overview of the study. (1)~Operational challenge: onboard EO imagery must be classified in orbit under strict compute, power, and downlink constraints, so that decisions, rather than raw imagery, are transmitted. (2)~Robustness-aware benchmark: following a train-on-ground, infer-onboard workflow, 14 pre-trained ViT and CNN backbones are compared on EuroSAT and PatternNet, on clean inputs and under realistic degradations (Gaussian noise, motion blur, and a DVB-S2(X)+JPEG transmission chain), across accuracy, efficiency, robustness, and ablation dimensions. (3)~Outcome: EfficientViT-M2 offers the best overall trade-off: near-top mean accuracy (98.54\%) at the lowest inference power (29.04~W) among the pre-trained ViTs, with compact compute (203.53~M MACs, 38.19~MB) and the strongest corruption robustness (overall score 0.79) with graceful SatCom degradation.}
    \label{fig:overview}
\end{figure*}

The computational cost of ViTs, driven by large parameter counts and attention, is a primary barrier to onboard deployment. RepSViT addresses this with a compact ViT (3.77\,M parameters, 0.6\,GFLOPs) suited to satellite missions, though such models still need further compression and optimization for complex tasks under tight energy budgets \cite{pang2024repsvit}. Neuromorphic processors offer another route by converting NNs into Spiking NNs, but this conversion is currently best supported for CNNs \cite{pang2024repsvit, lagunas2024performance, ortiz2024energy}, which, despite their efficiency, tend to trail Transformer models in both accuracy and corruption robustness on RS imagery benchmarks.

Transfer learning helps close this gap: fine-tuning a pre-trained model reuses rich hierarchical features from large-scale datasets, enabling rapid adaptation with far fewer labeled samples \cite{zhang2024vision}. This makes it especially valuable for resource-constrained OSS, where data and compute are limited. Beyond sample efficiency, there is a robustness rationale: representations learned from large, diverse corpora tend to be less tied to the idiosyncrasies of any single training set, so a fine-tuned pre-trained backbone can be expected to tolerate input degradations better than an equally sized model trained from scratch on clean RS data alone \cite{raghu2021vision, zhang2024vision}. Whether this expectation holds for compact ViTs under satellite-representative corruptions is precisely what our robustness evaluation tests.

Building on this strategy, this study identifies the most effective pre-trained model for land-use RS-IC in OSS, one that balances high accuracy, efficiency under limited compute, and robustness to the noise common in satellite inference. We deliberately adopt a train-on-ground, infer-onboard workflow, for three reasons. First, fine-tuning on the ground exploits abundant compute and labeled data, whereas training onboard is infeasible within current satellite power and thermal budgets. Second, only the compact fine-tuned weights must then be uplinked to the spacecraft, rather than raw imagery being downlinked for ground processing. Third, once a model is deployed, it is its \emph{inference-time} behaviour (power draw, memory footprint, and resilience to degraded inputs) that determines mission utility, so these are the quantities a selection study must measure. This last point also explains why model selection cannot rely on clean-data benchmark accuracy alone: a backbone that tops a leaderboard may draw hundreds of watts at inference or degrade sharply under the sensor noise, platform-induced blur, and downlink losses that onboard operation actually entails. We therefore compare CNN-, ResNet-, and Transformer-based models, evaluating accuracy, computational efficiency, and corruption robustness \emph{jointly} rather than in isolation, with particular attention to pre-trained ViTs under noisy onboard inference.

Our findings show that pre-trained ViTs significantly outperform models trained from scratch in accuracy, precision, and recall, at lower computational cost and with greater robustness to noisy test data. MobileViTV2 \cite{mehta2022separable} and EfficientViT-M2 \cite{liu2023efficientvit} emerge as the top performers: MobileViTV2 leads on clean data, while EfficientViT-M2 is more resilient under noise. Balancing training efficiency, accuracy, and robustness, EfficientViT-M2 is identified as the optimal choice for EO-OSS, offering a practical path to reliable, efficient onboard RS-IC. We summarize our contributions in translational AI for onboard satellite processing as follows:

\begin{itemize} 
\item \textbf{Energy-aware pretrained ViT selection:} A systematic exploration of pre-trained ViTs highlights EfficientViT-M2 as the best accuracy-per-power consumption option for onboard EO, satisfying the power budgets within which onboard processing has to operate.

\item \textbf{Comprehensive robustness validation:} Inference reliability is proven under unseen perturbations, across multiple benchmark EO datasets, and within a SatCom-in-the-loop framework that injects DVB-S2(X) channel impairments, demonstrating stable performance along the whole path from sensor capture to ground delivery.

\item \textbf{Like-for-like mission-context benchmarking:} All complexity figures (parameters, operation counts, and memory) are measured with one identical tool pipeline across every candidate and reconstructed flight-proven baselines (CloudScout of $\Phi$-Sat-1, RepSViT), yielding a self-consistent, hardware-agnostic edge-feasibility assessment rather than a cross-source comparison.
\end{itemize}

Fig.~\ref{fig:overview} summarizes the problem addressed, the experimental design, and the main result: among 14 pre-trained backbones evaluated under realistic onboard degradations, EfficientViT-M2 offers the best accuracy-per-watt and robustness trade-off for land-use classification performed onboard the spacecraft.

The remainder of this paper is organized as follows. Section~\ref{sec:rel_wrk} provides a brief survey of related works, serving as the basis for highlighting this paper's contributions. Section~\ref{sec:Mthdlg} presents the methodologies, including the dataset, machine learning (ML) models, and evaluation metrics employed in this comprehensive study. The experimental setup is explained in Section~\ref{sec:exp_setup}, followed by the numerical results and discussion in Section~\ref{sec:results}. Section~\ref{sec:limits} states the limitations of the study together with the future work they motivate, and Section~\ref{sec:concl} concludes the paper by summarizing the main findings and their implications for onboard deployment.

\vspace{-2mm}

\section{Related Works} \label{sec:rel_wrk}

Research relevant to this study falls into three strands: CNN enhancements for RS-IC, transfer-learning and pre-training strategies for bridging the natural-to-RS domain gap, and efficiency- and data-oriented innovations. We review each in turn, then position our contribution against recent architectures and domain-adaptation methods.

Within the first strand, recent RS work has enhanced CNNs through transfer learning, structural modifications, and attention. For example, \cite{albarakati2024novel} fused self-attention into a CNN for hyperspectral land-cover classification with custom augmentation, and \cite{rubab2024novel} combined dense blocks with inverted bottleneck residuals for stronger feature extraction. Closer to the onboard setting, our own ResNet-GLUSE \cite{le2026gluse} augments a compact ResNet with gated channel-wise (GLU-enhanced squeeze-and-excitation) attention and knowledge distillation from ViT teachers, reaching 94.63\% on EuroSAT and 98.09\% on PatternNet with only about 132K parameters. These designs confirm that attention and richer connectivity benefit RS features, but they retrofit such mechanisms onto CNN backbones rather than exploiting architectures pre-trained with attention from the outset; the present study instead asks how far compact pre-trained ViTs themselves can go under the same onboard power and memory constraints.

In the second strand, a key benefit of transfer learning is bridging the domain gap between natural and RS images, a persistent challenge given the absence of an ImageNet-scale RS benchmark. MGC \cite{li2024mgc} uses an MLP to guide CNN pre-training on small RS datasets, steering branches toward foreground regions for more discriminative features, while \cite{zhao2017transfer} pre-trains both feature descriptor and classifier of a deep CNN for high-resolution land-use classification, improving convergence and accuracy. The consistent gains reported by both justify our own protocol of fine-tuning ImageNet-pretrained backbones rather than training RS models from scratch.

Because ImageNet pre-training transfers imperfectly to RS, researchers have explored alternatives: MLPs mitigate distribution shift in cross-domain few-shot settings \cite{bai2024improving}; hybrid Transformer--CNN models process EuroSAT effectively \cite{depoian2023land}; and data-centric approaches leverage synthetic imagery \cite{fan2024scaling} or combine self- and supervised pre-training across RS and natural images, as in GeRSP \cite{huang2024generic}.

In the third strand, data scarcity and efficiency have driven further innovations: GeoSystemNet applies geosystem-aware DL to overcome labeled-data scarcity in high-resolution RS \cite{yamashkin2020improving}, while \cite{yu2024boosting} expands pre-trained models like CLIP with Mixture-of-Experts adapters for parameter-efficient continual learning. Interpretability has also gained attention; for instance, \cite{temenos2023interpretable} presents a SHAP-based framework revealing how spectral bands drive land-use predictions. Overall, while CNNs still dominate RS-IC, the uptake of ViTs, transfer learning, and synthetic data marks a clear shift toward more robust and broadly generalizable RS models.

Beyond the backbones benchmarked in this study, several recent architectures target RS classification. Tri-CNN \cite{alkhatib2023tricnn} combines multi-scale convolutional feature extraction for hyperspectral image classification, while SimPoolFormer \cite{alkhatib2025simpoolformer} adopts a simple-pooling transformer design to reduce attention cost for hyperspectral data. Both, however, are developed and validated on hyperspectral cubes with tens to hundreds of spectral bands; our study instead targets lightweight three-channel RGB classification with ImageNet-pretrained backbones suited to the onboard footprint, so these models are discussed as complementary context rather than directly benchmarked, to avoid an unfair cross-modality comparison. In a related direction, unsupervised domain adaptation (UDA) has been proposed to mitigate distribution shift in RS scene classification: \cite{binujose2026uda} combine adaptive incremental density-based clustering with multi-objective optimization, and \cite{li2025classaware} propose a class-aware UDA framework for cross-continental crop classification from Sentinel-2 time series. These works address cross-domain transfer with unlabeled target data, a setting orthogonal to the single-domain supervised classification studied here, and we identify sensor- and geography-induced domain shift as an important direction for future onboard-EO work.

Despite this rapid progress, a comprehensive comparison of downscaled, resource-efficient ViTs for onboard processing is still lacking. In particular, three operationally decisive questions remain open. First, existing RS-IC studies overwhelmingly report clean-data accuracy, so it is unknown whether the advantage of pre-trained ViTs survives the corruptions of onboard acquisition (sensor noise, platform-induced motion blur) and of the downlink itself (channel errors and lossy compression). Second, efficiency is usually summarized by parameter or FLOP counts, whereas the binding constraints of OSS are measured inference power and memory footprint, quantities that do not follow directly from FLOPs across heterogeneous architectures. Third, published complexity figures for flight-proven onboard models are sparse and measured under inconsistent conventions, preventing a fair answer to the practical question of whether a modern compact ViT is actually affordable relative to what has already flown. Our study addresses all three by jointly evaluating 14 backbones across accuracy, complexity, measured power, corruption robustness, and an end-to-end DVB-S2(X) transmission chain, under one consistent measurement protocol, providing practical model-selection guidance for onboard satellite EO.

\vspace{-2mm}

\section{Methodology} \label{sec:Mthdlg}

This section presents our methodological framework: the dataset and preprocessing, the ML models evaluated for RS-IC, and the metrics used to assess accuracy, computational efficiency, and robustness under the noisy conditions typical of real onboard satellite inference.

\vspace{-2mm}

\subsection{Dataset}
We use the EuroSAT benchmark \cite{helber2019eurosat}, derived from Sentinel-2 imagery: 27,000 geo-referenced $64{\times}64$ images with 13 spectral bands across 10 land-use classes (e.g., industrial and residential buildings, annual and permanent crops, rivers, seas/lakes, herbaceous vegetation, highways, pastures, forests). Its compact size and class diversity make it well suited to developing DL models for onboard EO. We additionally test on PatternNet \cite{zhou2018patternnet}, a retrieval-oriented set of 30,400 high-resolution $256{\times}256$ images over 38 categories, whose fine-grained diversity stresses the models on more complex RS scenes relevant to applications such as real-time environmental monitoring and precision agriculture.

Although EuroSAT provides 13 spectral bands, we deliberately use only the three RGB bands in this study. This choice keeps the input compatible with the three-channel ImageNet-pretrained backbones under comparison, allowing transfer learning without re-training the input stem; it matches the RGB configuration of the state-of-the-art baselines we benchmark against, ensuring a fair comparison; and it minimizes the onboard data volume and memory footprint. A consequence of discarding the near-infrared (NIR) band is reduced separability of spectrally similar classes, which we analyze against our confusion matrices in Section~\ref{sec:results}. A full NIR-inclusive multispectral study, which requires multispectral-native backbones, is identified as a direction for future work.

\vspace{-2mm}

\subsection{Machine Learning Models}

CNNs have long been the benchmark for visual data processing, but ViTs now achieve comparable or superior IC performance \cite{raghu2021vision}. We therefore evaluate a diverse set of models, trained from scratch and pre-trained, ranging from traditional CNNs to ViTs and convolution--Transformer hybrids, to compare lightweight, efficient models against more complex, high-capacity networks.

\subsubsection{Training from Scratch}
\begin{itemize}
    \item \textbf{CNN:} A basic CNN designed for IC, consisting of two convolutional layers followed by batch normalization, ReLU activation, and max pooling, with fully connected layers at the end for classification.
    \item \textbf{ResNet-14:} A smaller version of the ResNet framework, termed ResNet-14, with 14 layers, including two residual blocks in each of the three hidden layers, optimizing feature extraction and gradient flow through the network.
    \item \textbf{Compact Transformer (CCT) \cite{hassani2021escaping}:} A hybrid model that combines convolutional layers with transformer-based processing, using convolutional tokenization followed by transformer encoder layers for robust feature extraction and accurate image classification.
    \item \textbf{Small ViT \cite{lee2021vision}:} A small ViT model designed for efficiency, using a small patch tokenization process and a lightweight transformer with self-attention mechanisms to classify images effectively.
\end{itemize}

\subsubsection{Pretrained ViT Models}
\begin{itemize}
    \item \textbf{EfficientViT-M2 \cite{liu2023efficientvit}:} An efficient ViT combining convolutional layers with local window attention mechanisms optimized for balancing performance and computational efficiency with approximately 4 million parameters.
    \item \textbf{MobileViTV2 \cite{mehta2022separable}:} A hybrid model that combines convolutional and transformer-based processing, using depthwise separable convolutions and self-attention mechanisms for accurate and efficient IC.
    \item \textbf{xLSTM \cite{alkin2024vision}:} Integrates convolutional patch embedding with advanced LSTM-based layers to capture spatial and sequential dependencies for comprehensive IC.
    \item \textbf{EfficientNet-B2 \cite{tan2019efficientnet}:} A highly efficient CNN using depthwise separable convolutions and squeeze-and-excitation layers to minimize parameter usage while maintaining strong predictive performance.
    \item \textbf{ResNet50-DINO \cite{goldblum2024battle}:} A ResNet-50 model trained using self-supervised learning with DINO, optimized for feature representation without labeled data, enabling robust downstream classification performance.
    \item \textbf{EfficientViT-L2 \cite{cai2022efficientvit}:} Combines convolutional and transformer-based architectures with techniques like fused MBConv and Lite Multi-Head Attention to balance efficiency and representational power effectively.
    \item \textbf{SwinTransformer \cite{liu2021swin}:} A hierarchical ViT that divides images into non-overlapping patches, using shifted window-based self-attention to capture both local and global contextual information.
    \item \textbf{ViT \cite{DosovitskiyB0WZ21}:} A transformer model that divides images into patches, processing them through multiple transformer layers with self-attention to achieve high-capacity IC.
\end{itemize}

\begin{table*}[!t]
\centering
\footnotesize
\captionsetup{font=small}
\caption{\textsc{Qualitative Comparison of the Evaluated Machine Learning Models}}
\label{tab:model_comparison}
\begin{tabular}{@{}l p{4.6cm} p{4.9cm} p{4.9cm}@{}}
\toprule
\textbf{Model} & \textbf{Key design idea} & \textbf{Strengths} & \textbf{Limitations} \\
\midrule
\multicolumn{4}{@{}l}{\itshape Trained from scratch} \\
\midrule
CNN & Two convolutional layers with batch normalization and max pooling, fully connected head & Simplest architecture; lowest computational cost; fast to train & Learns only low-level features; weakest accuracy; prone to overfitting on small data \\
\addlinespace[2pt]
ResNet-14 & Compact residual network (two residual blocks per stage) & Stable gradient flow; strong accuracy per unit compute & Limited capacity for complex, fine-grained scenes \\
\addlinespace[2pt]
CCT & Convolutional tokenization followed by transformer encoder layers & Joint local--global context at small model size & Memorizes small datasets (large train--test gap); slower inference than plain CNNs \\
\addlinespace[2pt]
SmallViT & Lightweight patch tokenization with self-attention & Efficient pure-attention baseline & Limited representational capacity; longer training than CNN counterparts \\
\midrule
\multicolumn{4}{@{}l}{\itshape Pre-trained (fine-tuned)} \\
\midrule
EfficientViT-M2 & Convolutional stem with cascaded local-window attention ($\approx$4\,M params) & Best accuracy-per-watt; strongest corruption robustness; small footprint & Lower peak capacity than large ViTs \\
\addlinespace[2pt]
MobileViTV2 & Separable self-attention combined with depthwise convolutions & Highest clean-data accuracy; efficient hybrid design & Higher activation memory and inference power than EfficientViT-M2 \\
\addlinespace[2pt]
Vision-xLSTM & Convolutional patch embedding with xLSTM sequence blocks & Captures spatial and sequential dependencies & Extremely long training time; modest accuracy in this task \\
\addlinespace[2pt]
EfficientNet-B2 & Compound-scaled CNN with squeeze-and-excitation & High accuracy at very few FLOPs & Convolutional features only; large memory footprint \\
\addlinespace[2pt]
ResNet50-DINO & Self-supervised (DINO) pre-trained ResNet-50 & Label-free, transferable representations & Heavier than compact ViTs; mid-range accuracy \\
\addlinespace[2pt]
EfficientViT-L2 & Fused MBConv with lite multi-head attention & Scales well; strong accuracy on large data & Large size ($\approx$60\,M params); high memory demand \\
\addlinespace[2pt]
SwinTransformer & Hierarchical ViT with shifted-window attention & Near-top accuracy; effective multi-scale features & High compute, memory, and power cost \\
\addlinespace[2pt]
ViT (base/large/huge) & Global self-attention over image patches & Highest representational capacity & Data- and compute-hungry; diminishing returns and instability at the largest scale \\
\bottomrule
\end{tabular}
\end{table*}

Table \ref{tab:model_comparison} summarizes the key design idea, strengths, and limitations of each evaluated model. CNNs are ideal for simple, low-complexity tasks due to their ease of implementation and low computational cost, whereas ResNet-14 balances complexity and performance for mid-scale tasks. Compact Transformers and SmallViTs, while offering flexible and efficient architectures, may struggle with large datasets or higher complexity tasks. EfficientViT models, including M2 and L2, are well-suited for resource-constrained environments and high-performance tasks, respectively, but they require careful consideration of their computational limits. MobileViTV2 and Vision-xLSTM are powerful for specific tasks but demand significant computational resources. SwinTransformer and ViT models excel in capturing detailed and complex features, making them suitable for high-capacity IC tasks, though they come with high computational costs. Each model has its niche, making it ideal for different types of applications based on specific needs and resource availability.

\vspace{-2mm}

\subsection{Evaluation Metrics}

We evaluate multiclass performance over the $K{=}10$ classes with accuracy, precision, and recall \cite{goutte2005probabilistic}, computed per class and aggregated by support-weighted averaging:
\vspace{-2mm}

\begin{align}
&\text{Accuracy} = \frac{1}{N}\sum_{k=1}^K \mathrm{TP}_k, \nonumber \\
&\text{Precision} = \frac{1}{N} \sum_{k=1}^{K} N_k \frac{\mathrm{TP}_k}{\mathrm{TP}_k + \mathrm{FP}_k}, \nonumber \\
&\text{Recall} = \frac{1}{N} \sum_{k=1}^{K} N_k \frac{\mathrm{TP}_k}{\mathrm{TP}_k + \mathrm{FN}_k}, \nonumber
\end{align}
where $N$ is the total number of test samples, $N_k$ the number in class $k$, and $\mathrm{TP}_k$, $\mathrm{FP}_k$, and $\mathrm{FN}_k$ the true positives, false positives, and false negatives of class $k$. Weighting by the class support $N_k$ makes each aggregate representative of the full dataset; since EuroSAT classes are nearly balanced, weighted and macro averages practically coincide. These weighted metrics are used to select the best models in the final analysis.

To assess robustness, we follow the common-corruption benchmarking approach of \cite{HendrycksD19}. For a model $f$, let $A_{\text{clean}}$ and $A_{\phi}$ denote its accuracy on the clean test set and on the test set corrupted by perturbation $\phi$, respectively. The robustness score \cite{laugros2019adversarial} is
\begin{align}
 R^{\phi}_{f} = \frac{A_{\phi}}{A_{\text{clean}}},
\end{align}
where a score close to one indicates that the model largely retains its clean-data performance under the perturbation.

\vspace{-2mm}
 
\section{Experimental Setup} \label{sec:exp_setup}
This section describes the hardware and software configuration, the training and inference parameters, and the data augmentation and noise-simulation pipelines used to mimic real-world satellite EO conditions.

\vspace{-2mm}

\subsection{ML Model Configuration}
First, to ensure a balanced and comprehensive evaluation of the model's performance, the dataset was split into training and testing sets with a 70/30 ratio. 
For model optimization, different strategies have been employed based on the model architecture. The CNN and ResNet models were optimized using the Stochastic Gradient Descent (SGD) optimizer \cite{hardt2016train}, configured with a learning rate of 1e-3, momentum of 0.9, and a weight decay of 5e-4. In contrast, the ViT-based models, including the pre-trained variants, have been optimized using the AdamW optimizer \cite{loshchilovdecoupled}. This setup also used a learning rate 1e-3 and a weight decay 5e-4, ensuring consistency across different model types. A step learning rate scheduler was applied to further enhance training efficiency with a step size of 7 epochs and a decay factor ($\gamma$) of 0.1.

\begin{figure*}[!t]
    \centering
    \includegraphics[width=\linewidth]{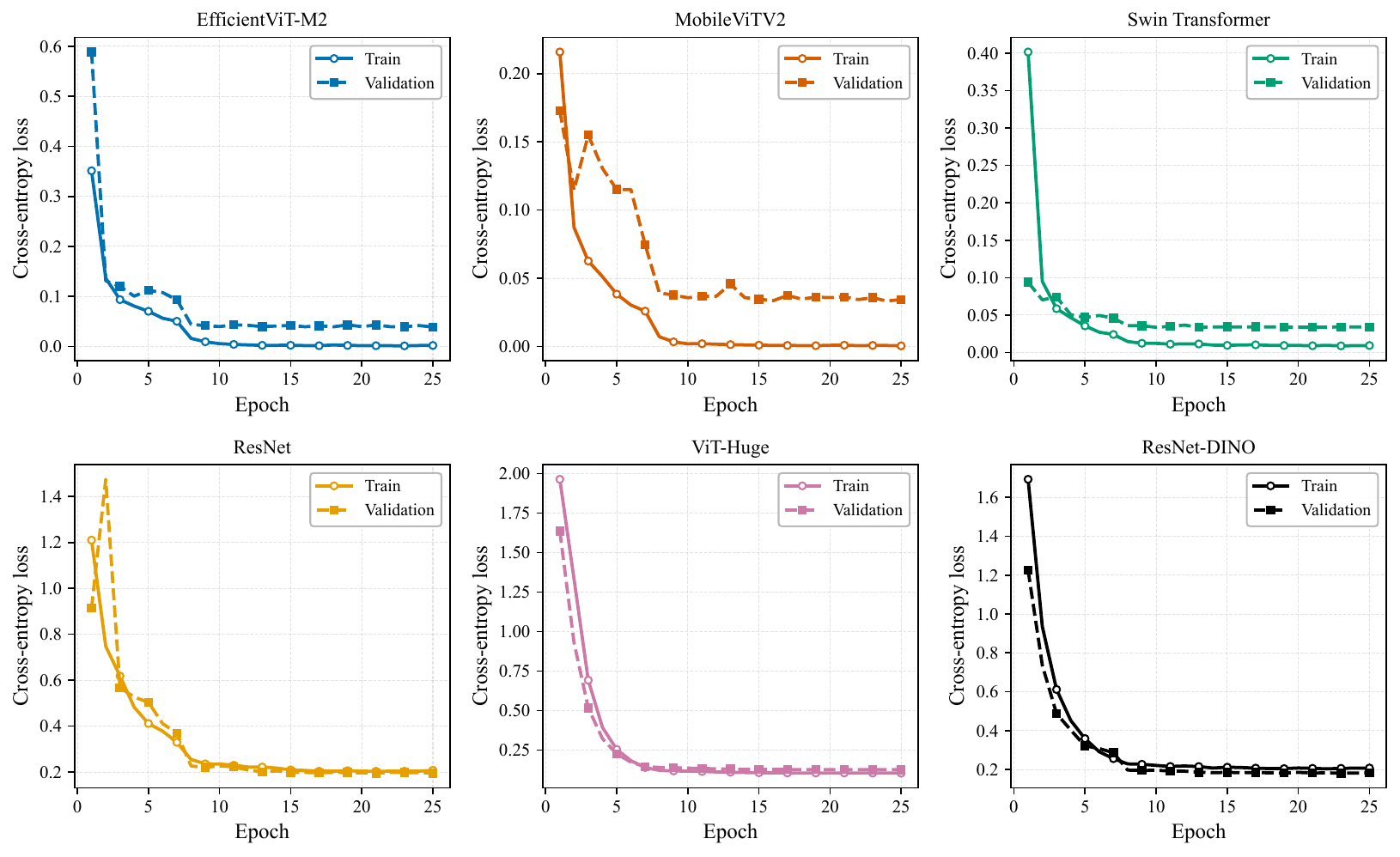}
    \captionsetup{font=small}
    \caption{Training (solid) and validation (dashed) loss versus epoch, one panel per backbone. Validation curves track training curves and plateau, evidencing no material over-fitting under the fixed 25-epoch schedule; compact ViT backbones settle at a lower loss floor.}
    \label{fig:per_model_loss}
    \vspace{-3mm}
\end{figure*}

All models were trained for a fixed budget of 25 epochs with no early stopping; an identical training budget avoids confounding the accuracy, robustness, and efficiency comparison with model-specific stopping criteria. Fig.~\ref{fig:per_model_loss} reports the training and validation loss versus epoch for six representative backbones, one panel per model. Every backbone shows a small generalization gap (final validation minus training loss $\leq 0.037$; ResNet and ResNet-DINO reach validation loss slightly below training loss), and each validation curve tracks its training curve and plateaus rather than diverging upward, so no material over-fitting occurs under the fixed schedule. The best-validation epoch, the natural stopping point an early-stopping criterion would select, is epoch~13 for Swin and epochs~21--25 for the other models, so early stopping would only reduce training cost without materially changing the reported accuracy. The compact ViT backbones (EfficientViT-M2, MobileViTV2, Swin) also settle at a markedly lower loss floor than the heavier ResNet, ResNet-DINO, and ViT-Huge baselines, consistent with the accuracy ranking in Section~\ref{sec:results}.

The CCT \cite{hassani2021escaping} and SmallViT models \cite{lee2021vision} have been trained using code adapted from the ViT-PyTorch repository, which provides reliable and optimized implementations for ViT. Other pre-trained models have been sourced from Huggingface's model hub or the official repositories in the respective papers. This approach ensured that the models were implemented consistently, with standardized architectures and preprocessing steps, allowing for a fair and direct comparison of their performance across the different experiments.\footnote{
\scriptsize
{Available codes and pre-trained models:\\
ViT-Pytorch: \url{https://github.com/lucidrains/vit-pytorch/tree/main}\\
EfficientViT-M2: \url{https://github.com/microsoft/Cream/tree/main/EfficientViT} \\
MobileViTV2: \url{https://huggingface.co/docs/transformers/en/model_doc/mobilevitv2} \\
Vision-xLSTM: \url{https://github.com/NX-AI/vision-lstm} \\
EfficientNet-b2: \url{https://huggingface.co/google/efficientnet-b2} \\
ResNet-DINO: \url{https://github.com/facebookresearch/dino} \\
EfficientViT-L2: \url{https://github.com/mit-han-lab/efficientvit} \\
SwinTransformer: \url{https://huggingface.co/docs/transformers/en/model_doc/swin} \\
ViT-base: \url{https://huggingface.co/google/vit-base-patch16-224} \\
ViT-large: \url{https://huggingface.co/google/vit-large-patch16-224} \\
ViT-huge: \url{https://huggingface.co/google/vit-huge-patch14-224-in21k}}
}

\vspace{-2mm}
\subsection{Hardware setup}
All experiments were conducted on a workstation with an Intel Xeon W-11855M CPU (12 cores, 3.20~GHz), 64~GB RAM, and an NVIDIA RTX A5000 GPU (48~GB GDDR6). The two largest pre-trained models, ViT-Large and ViT-Huge, required more extensive resources and were trained in parallel on a high-performance workstation with an Intel Xeon w9-3475X CPU (36 cores, up to 4.8~GHz), 512~GB DDR5 RAM, and two NVIDIA RTX 6000 Ada GPUs, each equipped with 48~GB of GDDR6 video memory.

\vspace{-2mm}
\subsection{Noisy Data for Inference}

\begin{figure*}[!ht]
    \centering
    \includegraphics[width=1\linewidth]{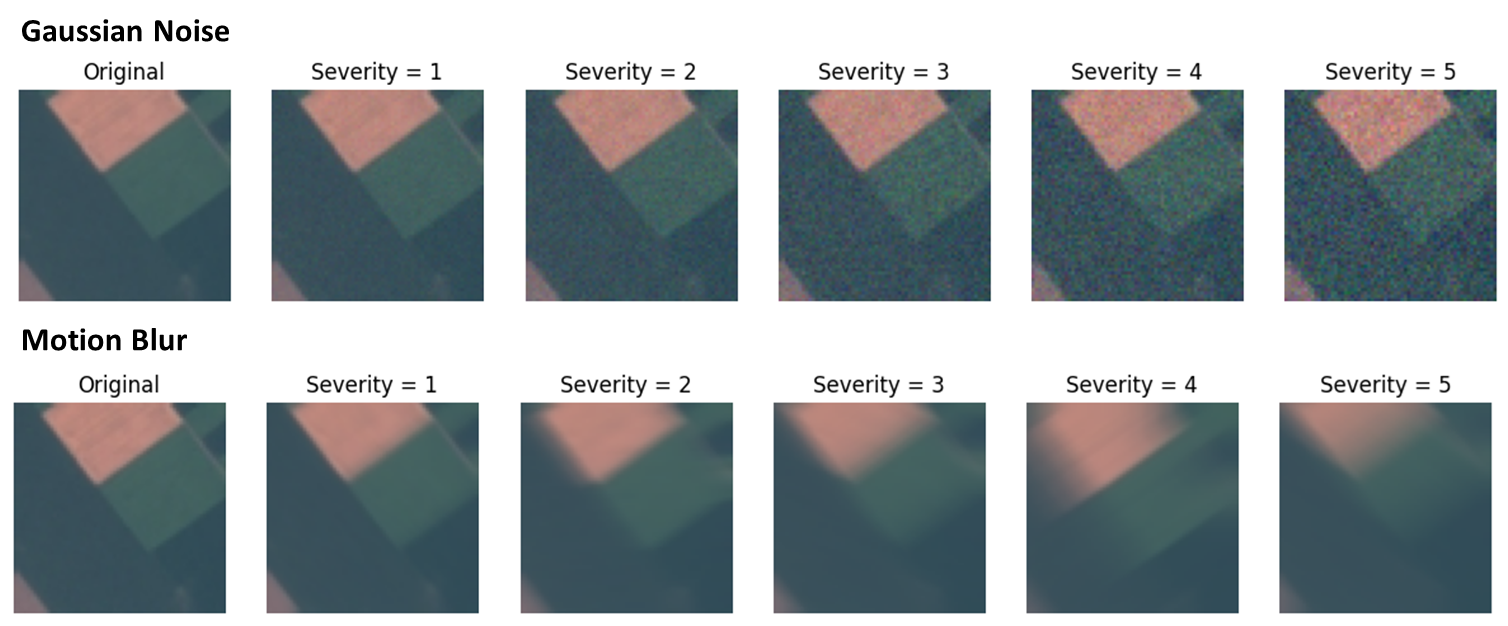}
    \captionsetup{font=small}
    \caption{Different noise levels with Gaussian and motion blur.}
    \label{fig:noise_level}
    \vspace{-2mm}
\end{figure*}

Although the models are trained on clean data, onboard inference data may be corrupted by natural events or instrument motion. We therefore assess robustness by injecting two corruption types into the inference test set, namely Gaussian noise and motion blur, each at five severity levels, as illustrated in Fig.~\ref{fig:noise_level}. The corruption pipeline follows the common-corruption benchmarking methodology of \cite{HendrycksD19}\footnote{\scriptsize https://github.com/hendrycks/robustness}, while the severity parameters are purpose-built for the onboard EO setting: rather than the heavier ImageNet-C constants, we adopt a low-to-moderate degradation regime representative of onboard acquisition conditions (sensor thermal noise and platform/attitude-induced motion). The exact mapping from the severity index to each corruption operator is defined below.

Each test image $\mathbf{x}\in[0,1]^{3\times64\times64}$ is corrupted at five discrete severity levels $s\in\{1,2,3,4,5\}$, where $s$ controls a single scalar parameter of each corruption operator such that a larger $s$ yields a strictly stronger corruption. For \emph{Gaussian noise}, additive, pixel-wise independent noise is applied in the normalized image domain and clipped to the valid range,
\begin{align}
\tilde{\mathbf{x}} &= \mathrm{clip}_{[0,1]}\!\big(\mathbf{x} + \mathbf{n}\big),
\quad
\mathbf{n}_{c,i,j} \stackrel{\text{i.i.d.}}{\sim}\mathcal{N}\!\big(0,\sigma^2(s)\big), \label{eq:gauss}\\
\sigma(s) &= 0.01\,s,
\qquad
\sigma \in \{0.01,\,0.02,\,0.03,\,0.04,\,0.05\}, \label{eq:gauss_sigma}
\end{align}
where $c$ indexes the RGB channel, $(i,j)$ the pixel location, and $\mathrm{clip}_{[0,1]}(\cdot)$ denotes clamping to $[0,1]$; the injected noise power thus grows quadratically with severity as $\sigma^2(s)=(0.01\,s)^2$. For \emph{motion blur}, the image is convolved with a linear, Gaussian-weighted kernel oriented at a random angle $\theta\sim\mathcal{U}(-45^\circ,45^\circ)$,
\begin{align}
\tilde{\mathbf{x}} = \mathbf{x} * K_{r,\sigma_b,\theta},
\qquad
K_{r,\sigma_b,\theta}(t) \propto
\exp\!\left(-\frac{t^2}{2\,\sigma_b^2(s)}\right), \label{eq:motion}
\end{align}
for $0\le t \le r(s)$, where $t$ is the signed distance (in pixels) from the kernel center along $\theta$, and the kernel is normalized to unit sum. The severity index jointly sets the kernel length (radius) $r(s)$ and the Gaussian spread $\sigma_b(s)$ according to Table~\ref{tab:severity}, both non-decreasing in $s$, so that larger $s$ corresponds to progressively longer and stronger directional averaging. All corruptions are applied only to the held-out test set; models are never trained on corrupted data.

\begin{table}[!t]
\centering
\captionsetup{font=small}
\caption{Corruption parameters as a function of the severity level $s$ for Gaussian noise and motion blur.}
\label{tab:severity}

\begin{tabular}{@{}l c c c c c@{}}
\toprule
\textbf{Severity $s$} & \textbf{1} & \textbf{2} & \textbf{3} & \textbf{4} & \textbf{5} \\
\midrule
Gaussian $\sigma(s)$              & 0.01 & 0.02 & 0.03 & 0.04 & 0.05 \\
Motion-blur radius $r(s)$         & 5    & 5    & 5    & 5    & 10   \\
Motion-blur spread $\sigma_b(s)$  & 1.5  & 2.0  & 3.0  & 4.0  & 5.0  \\
\bottomrule
\end{tabular}
\vspace{-2mm}
\end{table}

\begin{figure*}[!ht]
    \centering
    \includegraphics[width=1\linewidth]{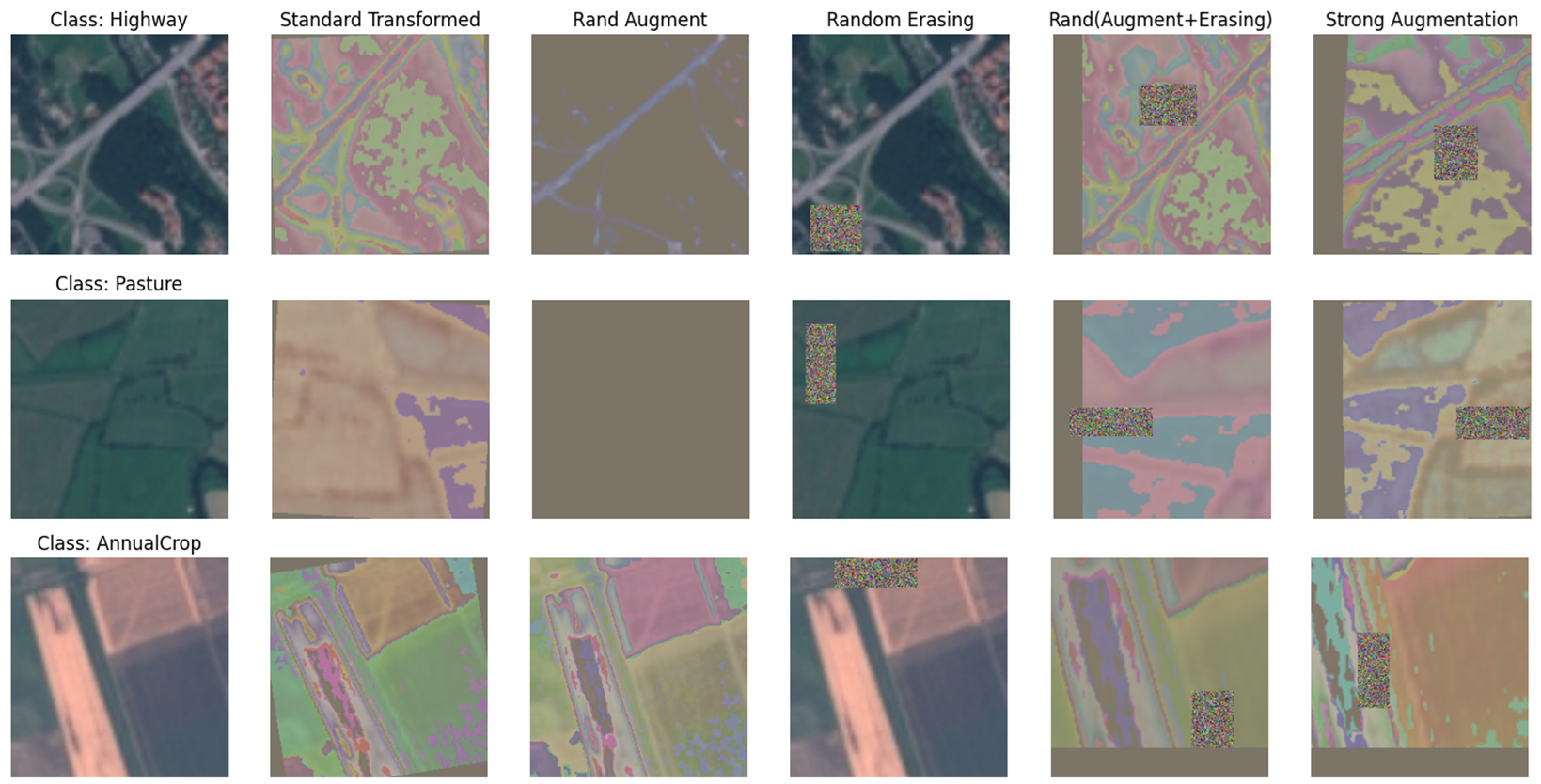}
    \captionsetup{font=small}
    \caption{Different augmentation techniques.}
    \label{fig:augment}
    \vspace{-2mm}
\end{figure*}

To further probe generalization, we also apply a series of data augmentation techniques (Fig.~\ref{fig:augment}): Standard Transform, RandAugment \cite{CubukZS020}, Random Erasing \cite{zhong2020random}, Rand (Augment+Erasing), and Strong Augmentation, which combines all of them. These introduce varying degrees of distortion and occlusion (color shifts, random erasures, and compound transformations), allowing us to test whether the models maintain performance on such altered inputs.

\section{Results and Discussions} \label{sec:results}

\begin{table*}[!ht]
\centering
\captionsetup{font=small}
\caption{\textsc{Summary of model computation complexity, estimated total size, training, and inference times}}
\vspace{-1mm}
\label{tab:model_complexity}
\begin{tabular}{@{}l c r r r r r@{}}
\toprule
\textbf{Models} & \textbf{Input Size} & \textbf{Total Parameters} & \textbf{Size (MB)} & \textbf{FLOPs} & \textbf{Training (s)} & \textbf{Inference (s)} \\
\midrule
CNN                           & 64$\times$64   & 66,330      & 0.71    & 0.93 MFLOPs   & 233     & 7   \\
ResNet                        & 64$\times$64   & 195,738     & 10.01   & 117.88 MFLOPs & 487     & 6.7 \\
CCT (Compact CNN-Transformer) & 64$\times$64   & 1,507,211   & 7.50    & 62.86 MFLOPs  & 241     & 23  \\
SmallViT                      & 64$\times$64   & 2,764,562   & 21.22   & 213.84 MFLOPs & 821     & 21  \\
EfficientViT-M2 (Microsoft)   & 224$\times$224 & 3,964,804   & 38.19   & 203.53 MFLOPs & 1000    & 10  \\
MobileViTV2 (Apple)           & 256$\times$256 & 4,393,971   & 259.30  & 1.84 GFLOPs   & 2,096   & 16  \\
Vision-xLSTM                  & 224$\times$224 & 6,090,098   & 141.66  & 1.86 GFLOPs   & 431,941 & 33  \\
EfficientNet-b2 (Google)      & 260$\times$260 & 7,715,084   & 252.86  & 32.92 MFLOPs  & 3,100   & 18  \\
ResNet-DINO (Facebook)        & 224$\times$224 & 23,528,522  & 272.54  & 4.14 GFLOPs   & 2,294   & 15  \\
EfficientViT-L2 (MiT)         & 224$\times$224 & 60,538,026  & 457.56  & 6.99 GFLOPs   & 6000    & 47  \\
SwinTransformer (Microsoft)   & 224$\times$224 & 86,753,474  & 636.38  & 15.47 GFLOPs  & 21,573  & 538 \\
ViT-base (Google)             & 224$\times$224 & 85,806,346  & 505.40  & 16.87 GFLOPs  & 8,582   & 46  \\
ViT-large (Google)$^{*}$      & 224$\times$224 & 303,311,882 & 1642.31 & 59.70 GFLOPs  & 9,271   & 60  \\
ViT-huge (Google)$^{*}$       & 224$\times$224 & 630,777,610 & 3454    & 162.00 GFLOPs & 13,533  & 75  \\
\bottomrule
\multicolumn{7}{@{}l}{\scriptsize $^{*}$Parallelly trained on a higher computing workstation with the following specifications:} \\
\multicolumn{7}{@{}l}{\scriptsize \textbf{CPU:} Intel Xeon w9-3475X (82.5 MB Cache, 36 cores, 72 threads, 4.8 GHz, 300 W), 512 GB DDR5 RAM, 4800 MHz} \\
\multicolumn{7}{@{}l}{\scriptsize \textbf{GPU:} Dual NVIDIA RTX 6000 Ada Generation, 48 GB GDDR6} \\
\end{tabular}
\vspace{-2mm}
\end{table*}

Table \ref{tab:model_complexity} compares the models' total parameters, estimated size, floating-point operations (FLOPs), and training and inference times. ViT-large and ViT-huge stand out for their substantial complexity, with high parameter counts, large memory footprints, and long training and inference times, whereas EfficientViT-M2 and MobileViTV2, though also ViT-family models, are far lighter across every metric. Within the Transformer family, complexity thus spans more than two orders of magnitude, and the compact variants offer a practical balance of performance and efficiency for deployment on resource-limited orbital platforms.

\begin{figure*}[!ht]
    \centering
    \includegraphics[width=0.9\linewidth]{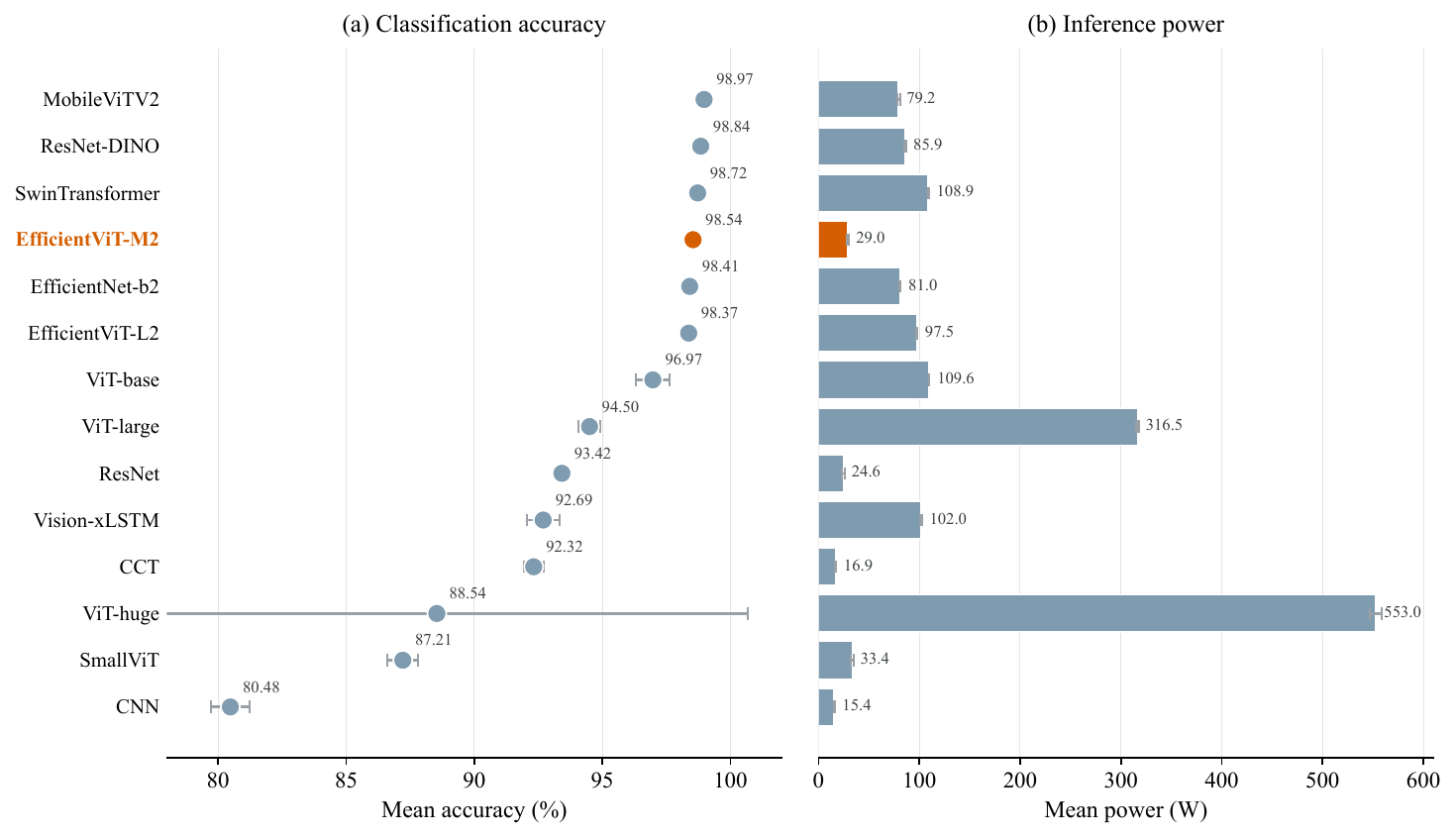}
    \vspace{-2mm}
    \captionsetup{font=small}
    \caption{Accuracy--efficiency comparison across models, sharing a common model axis ranked by mean top-1 accuracy: (a) mean top-1 accuracy and (b) mean inference power, each $\pm$ standard deviation over five runs on EuroSAT; EfficientViT-M2 highlighted. Mean precision and recall track accuracy to within $0.1$ percentage points for every model and are omitted from (a) for clarity.}
    \label{fig:perf_power}
    \vspace{-2mm}
\end{figure*} 

Based on five runs per model, Fig.~\ref{fig:perf_power} jointly characterizes accuracy (panel~a) and inference power (panel~b) on a shared, accuracy-ranked model axis. Panel~(a) confirms that fine-tuning pre-trained backbones consistently outperforms training from scratch, with stable accuracy, precision, and recall; MobileViTV2 leads (98.97\%), followed closely by SwinTransformer and the EfficientViT variants. Increasing model size does not guarantee better performance: ViT-huge attains only 88.54\% mean accuracy with large run-to-run variability, indicating overfitting or optimization difficulties in such complex architectures, so model size must be balanced against performance rather than assumed beneficial. Panel~(b) reveals the complementary efficiency picture: EfficientViT-M2 draws just 29.04$\pm$0.96~W, the lowest among all pre-trained ViT models and far below MobileViTV2 (79.23$\pm$1.45~W) and ViT-huge (552.96$\pm$6.19~W), while remaining close to lightweight from-scratch baselines such as CNN (15.42$\pm$0.52~W) and ResNet (24.63$\pm$1.63~W). Read together, the two panels show that EfficientViT-M2 sits near the accuracy frontier yet at a fraction of the power, making it especially attractive for energy-constrained onboard satellite processing where accuracy must be traded against a strict power budget.

\begin{table*}[!ht]
\centering
\captionsetup{font=small}
\caption{\textsc{Experimental results for the best model's performance on EuroSat}}
\vspace{-1mm}
\label{tab:experiment_compare}
\begin{tabular}{@{}l r r r r r r@{}}
\toprule
\textbf{Models} & \textbf{Train Loss} & \textbf{Test Loss} & \textbf{Train Accuracy} & \textbf{Accuracy} & \textbf{Precision} & \textbf{Recall} \\
\midrule
CNN                           & 0.469      & 0.534     & 83.73          & 81.82    & 81.07     & 81.19  \\
ResNet                        & 0.217      & 0.206     & 93.5           & 93.88    & 93.89     & 93.88  \\
CCT (Compact CNN-Transformer) & 0.038      & 0.26      & 99.02          & 92.61    & 92.7      & 92.61  \\
SmallViT                      & 0.25       & 0.42      & 90.98          & 86.49    & 86.9      & 86.49  \\
EfficientViT-M2 (Microsoft)   & 0.002      & 0.039     & \underline{\textit{99.99}}          & \underline{\textbf{98.76}}    & \underline{\textbf{98.77}}     & \underline{\textbf{98.76}}  \\
MobileViTV2 (Apple)           & 0.00036    & 0.034     & \textbf{100}            & \textbf{99.09}    & \textbf{99.09}     & \textbf{99.09}  \\
Vision-xLSTM                  & 0.34       & 0.41      & 86.9           & 85.8     & 85.9      & 85.8   \\
EfficientNet-b2 (Google)      & 0.007      & 0.07      & 99.9           & 98.47    & 98.49     & 98.47  \\
ResNet-DINO (Facebook)        & 0.28       & 0.18      & 94.2           & 94.9     & 94.5      & 94.9   \\
EfficientViT-L2 (MiT)         & 0.0006     & 0.081     & \underline{\textit{99.99}}          & 98.14    & 98.16     & 98.15  \\
SwinTransformer (Microsoft)   & 0.009      & 0.034     & 99.8           & \underline{\textit{98.83}}    & \underline{\textit{98.83}}     & \underline{\textit{98.83}}  \\
ViT-base (Google)             & 0.007      & 0.05      & 99.96          & 98.45    & 98.47     & 98.45  \\
ViT-large (Google)            & 0.002      & 0.04      & 100            & 98.55    & 98.55     & 98.55  \\
ViT-huge (Google)             & 0.105      & 0.127     & 97.5           & 96.39    & 96.39     & 96.39  \\
\bottomrule
\multicolumn{7}{@{}l}{\scriptsize \textbf{Bold} denotes the best values; \underline{\textit{italic and underline}} the second best; \underline{\textbf{bold and underline}} the third best.} \\
\end{tabular}
\vspace{-6mm}
\end{table*}

Considering the best run per model (Table \ref{tab:experiment_compare}), MobileViTV2, SwinTransformer, and EfficientViT-M2 lead: MobileViTV2 reaches 99.09\% test accuracy, precision, and recall, followed by SwinTransformer (98.83\%) and EfficientViT-M2 (98.76\%). Among the large pre-trained models, SwinTransformer outperforms ViT-base, ViT-large, and ViT-huge despite having the fewest parameters (86~M); its hierarchical, shifted-window attention captures local and global features more effectively than global attention over the whole image, confirming that architectural design, rather than raw parameter count, drives this performance advantage.

Together, Tables \ref{tab:model_complexity} and \ref{tab:experiment_compare} and Fig.~\ref{fig:perf_power} identify MobileViTV2 and EfficientViT-M2 as the two best models for EuroSAT classification, combining low computational complexity and short training/inference times with top-tier metrics. Against the state of the art (Table \ref{tab:compare_sota}), MobileViTV2 achieves the highest scores and EfficientViT-M2 the second highest, confirming both as leading choices among existing methods.

\begin{figure}[!ht]
    \centering
    \includegraphics[width=\linewidth]{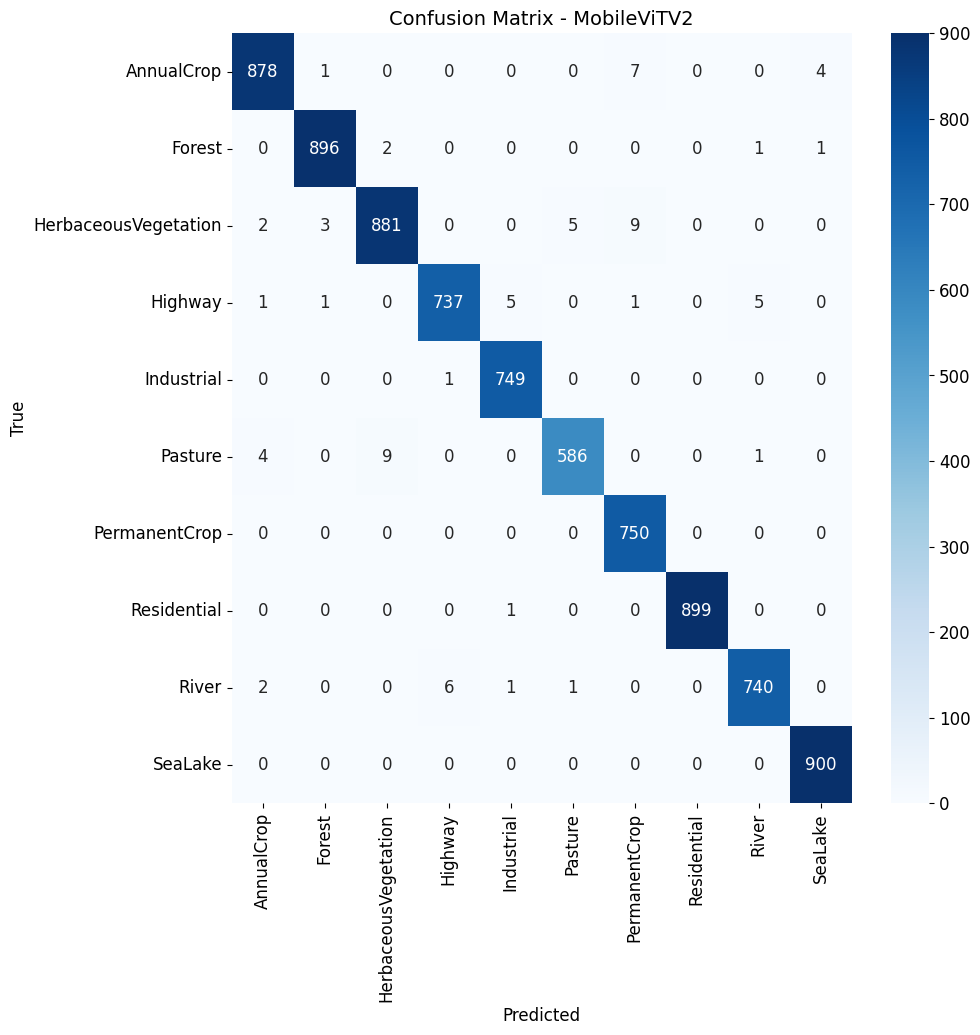}
    \includegraphics[width=\linewidth]{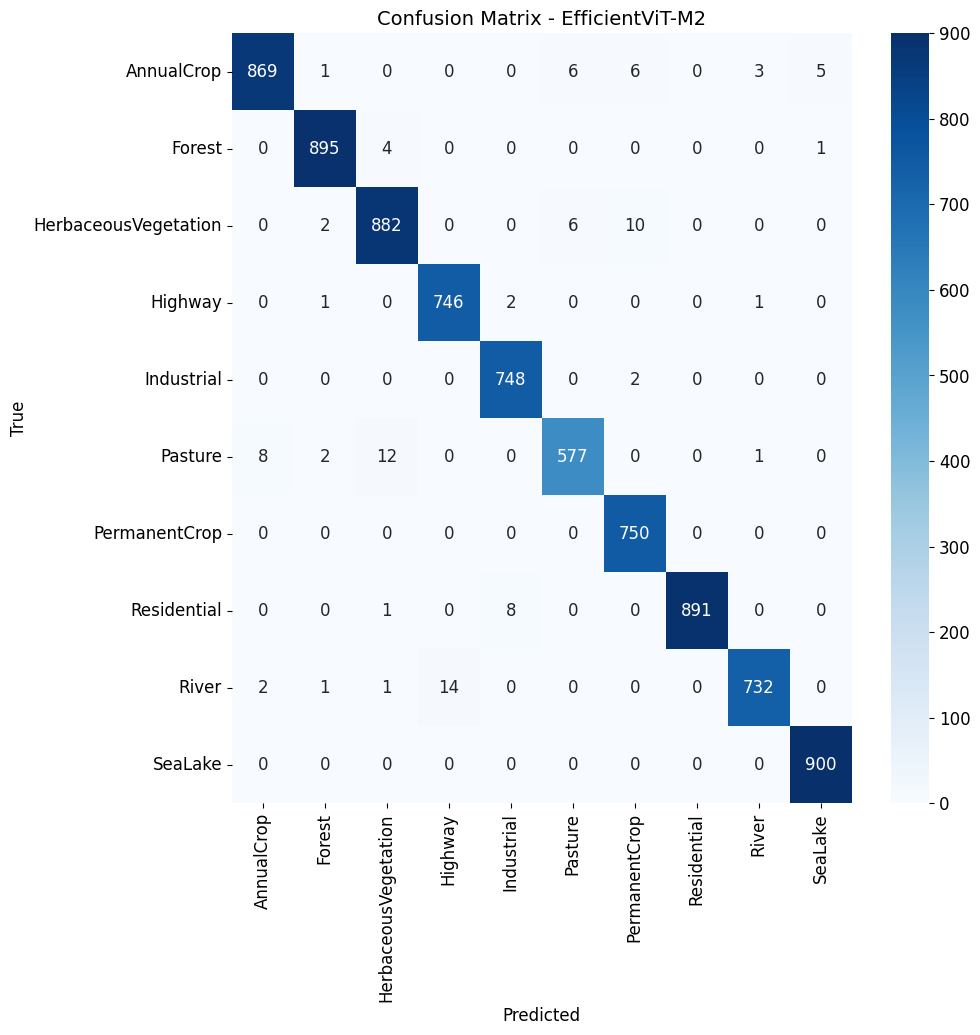}
    \vspace{-3mm}
    \captionsetup{font=small}
    \caption{Confusion matrix from MobileViTV2 (top) and EfficientViT-M2 (bottom) performance on EuroSat.}
    \label{fig:mobileViT_cf}
    \vspace{-2mm}
\end{figure}

The confusion matrices in Fig.~\ref{fig:mobileViT_cf} confirm this picture: both models classify most classes almost perfectly (e.g., \textit{Forest}, \textit{PermanentCrop}, \textit{SeaLake}), with only minor residual errors concentrated in \textit{Pasture} and \textit{River}, and MobileViTV2 makes slightly fewer misclassifications overall.

\begin{table}[!t]
\centering
   \captionsetup{font=small}
\caption{\textsc{Performance comparison with the-state-of-the-art models (Top-1) on EuroSat.}}
 \vspace{-1mm}
\label{tab:compare_sota}
\begin{tabular}{@{}l c c c@{}}
\toprule
\textbf{Models} & \textbf{Accuracy} & \textbf{Precision} & \textbf{Recall} \\
\midrule
EfficientViT-M2                          & \underline{\textit{98.76}}   & \underline{\textit{98.77}}     & \underline{\textit{98.76}}  \\
MobileViTV2                              & \textbf{99.09}    & \textbf{99.09}     & \textbf{99.09}  \\
Few-shot MLP \cite{bai2024improving}     & 79.62    & --        & --     \\
Attention+CNN \cite{albarakati2024novel} & 89.5     & --        & --     \\
ACL \cite{xu2024attention}               & 95.46    & --        & --     \\
MGC \cite{li2024mgc}                     & 96.41    & --        & --     \\
GeRSP \cite{huang2024generic}            & 97.87    & --        & --     \\
Constrastive Learning \cite{fan2024scaling} & 96    & --        & --     \\
GeoSystemNet \cite{yamashkin2020improving} & 95.32  & --        & --     \\
Transformer+CNN \cite{depoian2023land}   & 95.48    & --        & --     \\
MoE-ViT \cite{yu2024boosting}            & 98.1     & --        & --     \\
SIBNet \cite{rubab2024novel}            & 97.8     & 97        & 96.97  \\
CNN-SHAP \cite{temenos2023interpretable} & 94.72    & 93.73     & 94.10  \\
CNN-SVM \cite{tumpa2024lightweight}      & 97.91    & --        & --     \\
\bottomrule
\multicolumn{4}{@{}p{0.97\columnwidth}@{}}{\scriptsize \textbf{Bold} denotes the best values and \underline{\textit{italic and underline}} the second best; ``--'' indicates that the cited work does not report the corresponding precision or recall value.} \\
\end{tabular}
\vspace{-2mm}
\end{table}

\begin{figure*}[!ht]
    \centering
    \includegraphics[width=\linewidth]{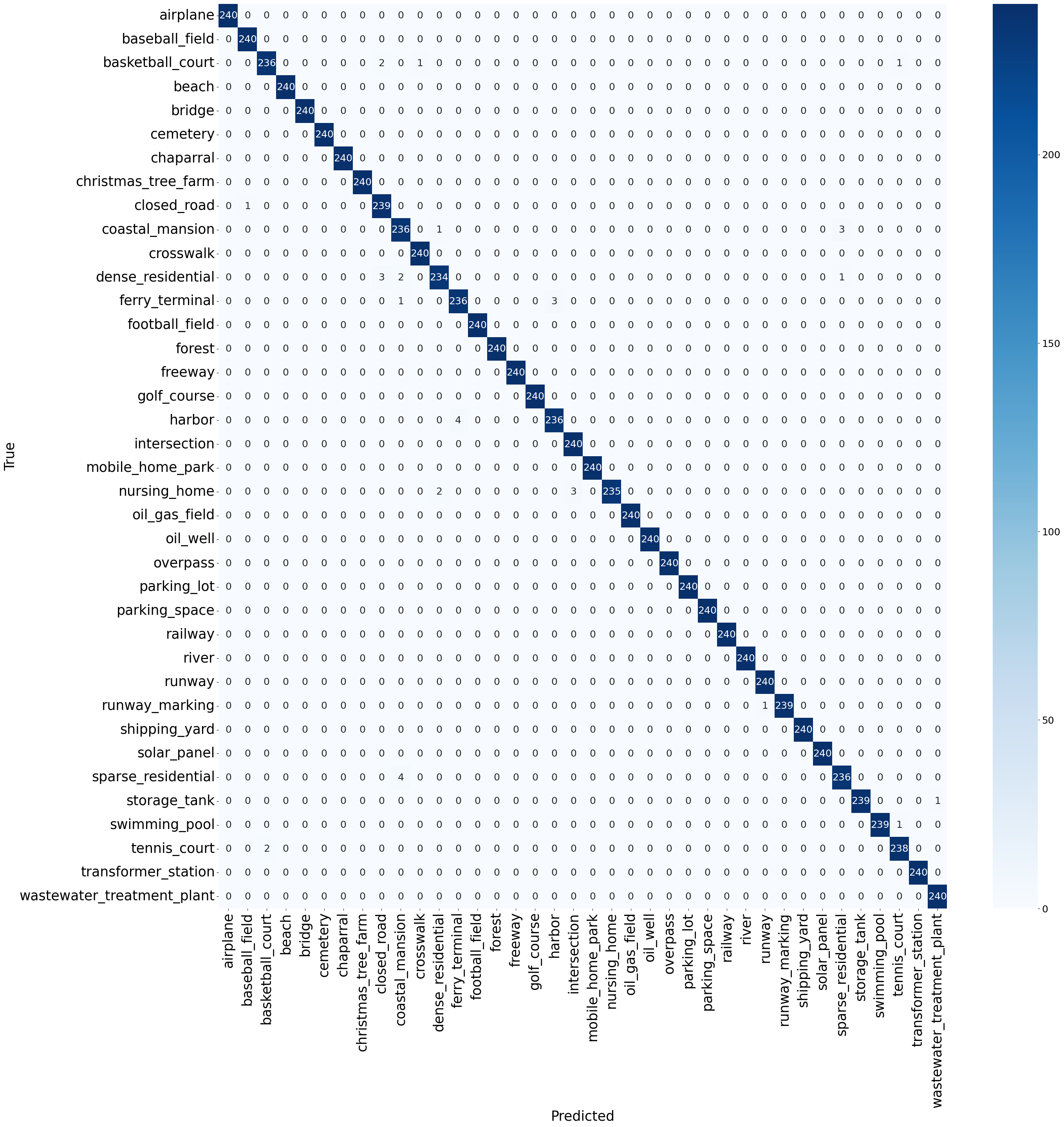}
    \captionsetup{font=small}
    \caption{Confusion matrix of EfficientViT-M2 on PatternNet.}
    \label{fig:conf_EfficientViT_PatternNet}
    \vspace{-2mm}
\end{figure*}

On PatternNet, Table \ref{tab:compare_sota_PatternNet} and the confusion matrix in Fig.~\ref{fig:conf_EfficientViT_PatternNet} show near-perfect scene recognition: EfficientViT-M2 attains 99.52\% Top-1 accuracy, precision, and recall, within 0.14 points of the best-reported score and above SwinTransformer (99.23\%) and ResNet-50 (98.23\%) \cite{peng2020efficient}, with consistent predictions across all 38 land-use categories, reaffirming its suitability for onboard EO where predictive performance and resource efficiency are both critical.

\begin{table}[!t]
\centering
   \captionsetup{font=small}
\caption{\textsc{Performance comparison with the-state-of-the-art models (Top-1) on PatternNet.}}
 \vspace{-1mm}
\label{tab:compare_sota_PatternNet}
\begin{tabular}{@{}l c c c@{}}
\toprule
\textbf{Models} & \textbf{Accuracy} & \textbf{Precision} & \textbf{Recall} \\
\midrule
EfficientViT-M2                     & \underline{\textbf{99.52}}   & \underline{\textbf{99.52}}     & \underline{\textbf{99.52}}  \\
MobileViTV2                         & \textbf{99.66}    & \textbf{99.66}     & \textbf{99.66}  \\
EfficientViT-b2                     & \underline{\textit{99.54}}   & \underline{\textit{99.54}}     & \underline{\textit{99.53}}  \\
ResNet-DINO                         & 96.01    & 96.01     & 96.00  \\
SwinTransformer                     & 99.23    & 99.23     & 99.23  \\
ResNet-50 \cite{peng2020efficient}  & 98.23    & 97.95     & --     \\
\bottomrule
\multicolumn{4}{@{}p{0.97\columnwidth}@{}}{\scriptsize \textbf{Bold} denotes the best values, \underline{\textit{italic and underline}} the second best, and \underline{\textbf{bold and underline}} the third best; ``--'' indicates that the cited work does not report the corresponding recall value for that model.} \\
\end{tabular}
\vspace{-2mm}
\end{table}

The confusion matrices in Fig.~\ref{fig:mobileViT_cf} also clarify the effect of the RGB-only band choice discussed in Section~\ref{sec:Mthdlg}. The dominant residual errors concentrate on spectrally similar, RGB-ambiguous class pairs, most notably River versus Highway, and among vegetation-type classes (e.g., Pasture and Herbaceous Vegetation). These are precisely the confusions that near-infrared (NIR)-derived indices are designed to resolve: the Normalized Difference Water Index, $\mathrm{NDWI}=(G-\mathrm{NIR})/(G+\mathrm{NIR})$, would separate the elongated dark water bodies (rivers) from paved surfaces (highways) that appear similar under RGB, while the Normalized Difference Vegetation Index, $\mathrm{NDVI}=(\mathrm{NIR}-R)/(\mathrm{NIR}+R)$, would sharpen boundaries among vegetation classes. This indicates that a multispectral extension is a promising route to reduce these specific errors, at the cost of the additional bandwidth, memory, and multispectral-native backbones that the onboard-efficiency focus of this study deliberately avoids; we therefore leave this direction to future work.

\begin{table}[!ht]
\centering
\captionsetup{font=small}
\caption{\textsc{MobileViTV2 Performance with Different Augmentation Techniques}}
 \vspace{-1mm}
\label{tab:mobile_aug}
\begin{tabular}{@{}l c c c@{}}
\toprule
\textbf{MobileViTV2} & \textbf{Accuracy} & \textbf{Precision} & \textbf{Recall} \\
\midrule
Baseline        & \textbf{99.09}    & \textbf{99.09}     & \textbf{99.09}  \\
RandAugment     & 98.68    & 98.68     & 98.68  \\
StandardAugment & 98.23    & 98.25     & 98.23  \\
RandErasing     & 98.73    & 98.73     & 98.73  \\
Rand(Aug+Eras)  & 98.39    & 98.42     & 98.39  \\
StrongAugment   & 98.24    & 98.25     & 98.24  \\
\bottomrule
\multicolumn{4}{@{}l}{\scriptsize \textbf{Bold} denotes the best values.} \\
\end{tabular}
\vspace{-2mm}
\end{table}

\begin{table}[!ht]
\centering
\captionsetup{font=small}
\caption{\textsc{EfficientViT Performance with Different Augmentation Techniques}}
 \vspace{-1mm}
\label{tab:efficient_aug}
\begin{tabular}{@{}l c c c@{}}
\toprule
\textbf{EfficientViT-M2} & \textbf{Accuracy} & \textbf{Precision} & \textbf{Recall} \\
\midrule
Baseline        & \textbf{98.76}    & \textbf{98.77}     & \textbf{98.76}  \\
RandAugment     & 98.59    & 98.6      & 98.59  \\
StandardAugment & 97.39    & 97.44     & 97.39  \\
RandErasing     & 98.22    & 98.24     & 98.22  \\
Rand(Aug+Eras)  & 97.96    & 97.99     & 97.96  \\
StrongAugment   & 97.09    & 97.14     & 97.1   \\
\bottomrule
\multicolumn{4}{@{}l}{\scriptsize \textbf{Bold} denotes the best values.} \\
\end{tabular}
\vspace{-5mm}
\end{table}

\begin{figure*}[!ht]
    \centering
    \includegraphics[width=0.9\linewidth]{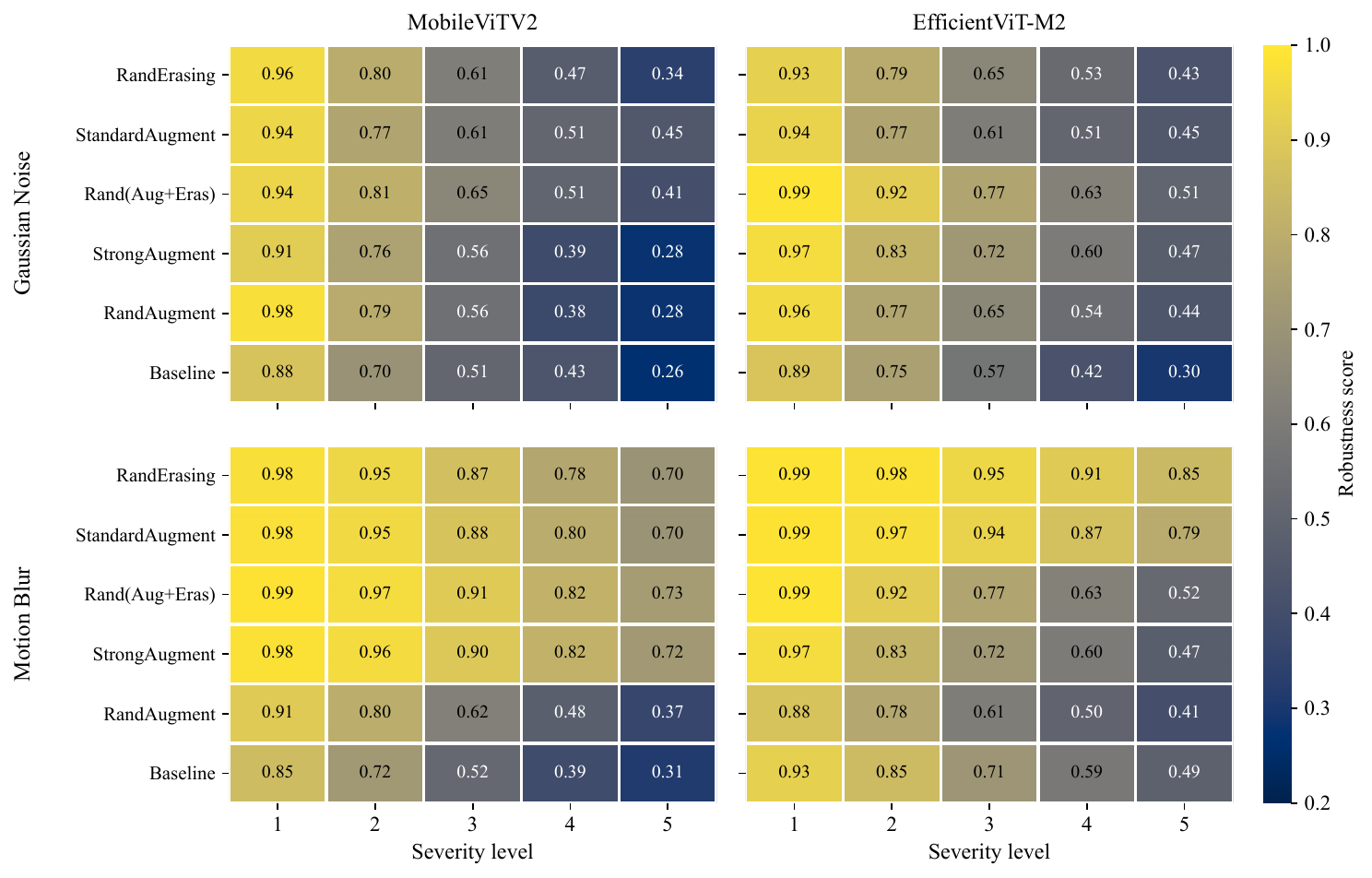}
    \captionsetup{font=small}
    \caption{Robustness of MobileViTV2 (left) and EfficientViT-M2 (right) under Gaussian noise (top row) and motion blur (bottom row). Each cell reports the robustness score of one augmentation technique (rows, sorted by overall mean) at one severity level (columns, $1$--$5$); a brighter cell indicates greater retention of classification performance under corruption, on a color scale shared by all four panels.}
    \label{fig:robustness}
    \vspace{-4mm}
\end{figure*}

Tables \ref{tab:mobile_aug} and \ref{tab:efficient_aug} show that training-time augmentation slightly reduces clean-data performance: the un-augmented baselines score highest, while RandErasing stays closest to baseline for both models (98.73\% for MobileViTV2, 98.22\% for EfficientViT-M2). RandErasing thus adds enough variability to enhance robustness at minimal clean-data cost, offering the best balance for real-world noisy inputs.

Under Gaussian noise (Fig.~\ref{fig:robustness}, top row), both models retain high robustness at low severity (levels 1--2), with EfficientViT-M2 slightly ahead (at level~1, 0.989 with Rand(Aug+Eras) versus MobileViTV2's best of 0.976 with RandAugment). As severity grows, robustness declines for both, but EfficientViT-M2 stays consistently higher at levels 3--5, where Rand(Aug+Eras) and StrongAugment are most effective: at level~5 it reaches 0.505 with Rand(Aug+Eras) versus MobileViTV2's best of 0.447 with StandardAugment. RandErasing gives MobileViTV2 its most balanced behaviour, but the model degrades faster than EfficientViT-M2 as the injected noise severity increases.

The motion-blur results (Fig.~\ref{fig:robustness}, bottom row) follow the same pattern. Both models are resilient at level~1 (EfficientViT-M2 up to 0.989 with RandErasing; MobileViTV2 up to 0.986 with Rand(Aug+Eras)), but at level~5 EfficientViT-M2 retains 0.846 with RandErasing against MobileViTV2's best of 0.735, and it maintains higher best-case scores across all severities. MobileViTV2 declines more sharply, especially with RandAugment and no augmentation; StrongAugment mitigates this at high severity but does not close the gap.

Overall, the Gaussian-noise and motion-blur experiments consistently demonstrate the superior corruption robustness of EfficientViT-M2, particularly at the high severities where performance is hardest to maintain, making it the more dependable choice for onboard satellite classification under realistically degraded imaging conditions.

The practicality of EfficientViT-M2 for OSS deployment is further supported by its compact footprint (38.19~MB, 203.53~MFLOPs) relative to models flown on pioneering missions: the CloudScout model used for onboard cloud detection on $\Phi$-Sat-1 \cite{giuffrida2021varphi} requires approximately 4.67~GFLOPs and 123.98~MB. Because the original CloudScout publication does not report parameter count, computational cost, or memory footprint, these figures were not taken from the literature; instead, we reconstructed the CloudScout architecture as described in \cite{giuffrida2021varphi} and measured it with the \emph{same} tools and input convention used for every model in this study: parameter and operation counts via \texttt{fvcore}\footnote{Computational complexity is reported as multiply--accumulate operations (MACs) measured with \texttt{fvcore}, applied identically to all models to ensure a fully self-consistent cross-model comparison.} and total (parameter~+~activation) memory via \texttt{torchinfo}, at a $3\times192\times192$ input. This yields 158{,}083 parameters, $\approx$4.67~G MACs, and a 123.98~MB estimated total size, the latter dominated by high-resolution activation memory rather than weights. Reporting all models through one identical measurement pipeline is what makes EfficientViT-M2's $\approx$23$\times$ lower computational cost a meaningful, like-for-like comparison rather than a cross-source one. Additionally, we restrict the comparison with the $\Phi$-Sat-2 mission \cite{guerrisi2023artificial} to model complexity only. That mission employed a convolutional autoencoder with 81.2 MFLOPs for an image-\emph{compression} task on an Intel Movidius Myriad 2 VPU, whereas EfficientViT-M2 performs image \emph{classification} and was timed on a GPU; because the two differ in task, workload, and hardware, we make no claim regarding relative speed or accuracy. On a complexity basis, EfficientViT-M2 remains lightweight and comparable in scale, and a fair latency/accuracy assessment would require a same-task, same-hardware benchmark, which we leave to future work. Finally, compared with RepSViT \cite{pang2024repsvit} (3.77~M parameters, 600~MFLOPs for onboard image processing), EfficientViT-M2 offers a $\approx$3$\times$ reduction in complexity, reinforcing its balance of performance and resource efficiency for real-time image processing onboard the spacecraft.

\vspace{-2mm}
\subsection{Ablation Study}

To consolidate the preceding analyses, we summarize the ablation evidence along three axes: (i) a \emph{data-augmentation} ablation (Tables~\ref{tab:mobile_aug} and \ref{tab:efficient_aug}), which isolates the effect of each augmentation policy on clean and corrupted accuracy; (ii) a \emph{pre-training} ablation, contrasting pretrained backbones against training from scratch (Fig.~\ref{fig:perf_power}(a)); and (iii) a new \emph{end-to-end DVB-S2(X) transmission} ablation, described next, which stresses the models under realistic downlink conditions.

\begin{figure}[!t]
    \centering
    \includegraphics[width=\linewidth]{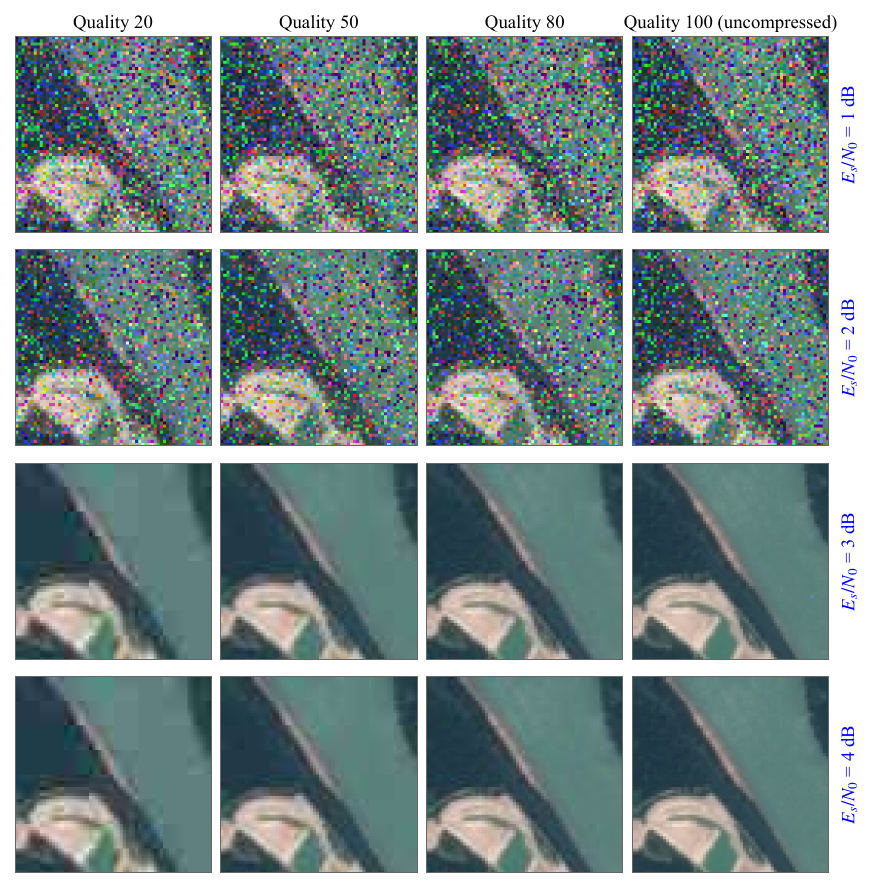}
    \captionsetup{font=small}
    \caption{Preparation of the DVB-S2(X) evaluation corpus, shown for one representative EuroSAT River scene. Columns vary the JPEG compression quality applied before transmission, and rows vary the channel condition $E_s/N_0$ of the DVB-S2(X) link. Each cell is the image as recovered after decoding, which is exactly what the classifiers receive at the corresponding operating point of Fig.~\ref{fig:dvbs2x}. Compression suppresses fine detail and creates block artifacts, while a low link margin scatters residual bit errors across the scene.}
    \label{fig:dvbs2x_samples}
    \vspace{-3mm}
\end{figure}

Before reporting the outcome of the third ablation, Fig.~\ref{fig:dvbs2x_samples} illustrates how its evaluation corpus was prepared. Every EuroSAT image is JPEG-compressed at a prescribed quality factor, transmitted over a DVB-S2(X) link at a prescribed channel SNR, and then decoded, so that each operating point of the ablation corresponds to one fully regenerated copy of the dataset rather than to a synthetic pixel-level perturbation. The emulation chain that generates this corpus follows the companion end-to-end SatCom study \cite{nguyen2025semantic}, whereas the comparative evaluation of onboard classifiers over the corpus, reported below, is specific to the present work. The figure arranges the two axes jointly for a single River scene: columns vary the compression quality at a fixed channel condition, and rows vary $E_s/N_0$ at a fixed quality. Two effects are visible. Compression alone removes fine radiometric detail and introduces block artifacts along the river bank, which is most apparent at quality~20 in the two lower rows, whereas an insufficient link margin injects residual bit errors that survive decoding and scatter impulsive color noise across the whole scene, leaving only the coarsest structures interpretable at 1 and 2~dB. The identical chain is applied to all ten EuroSAT classes and to both the training and test partitions, and the single scene shown here is representative of that corpus.

\begin{figure*}[!t]
    \centering
    \includegraphics[width=0.82\linewidth]{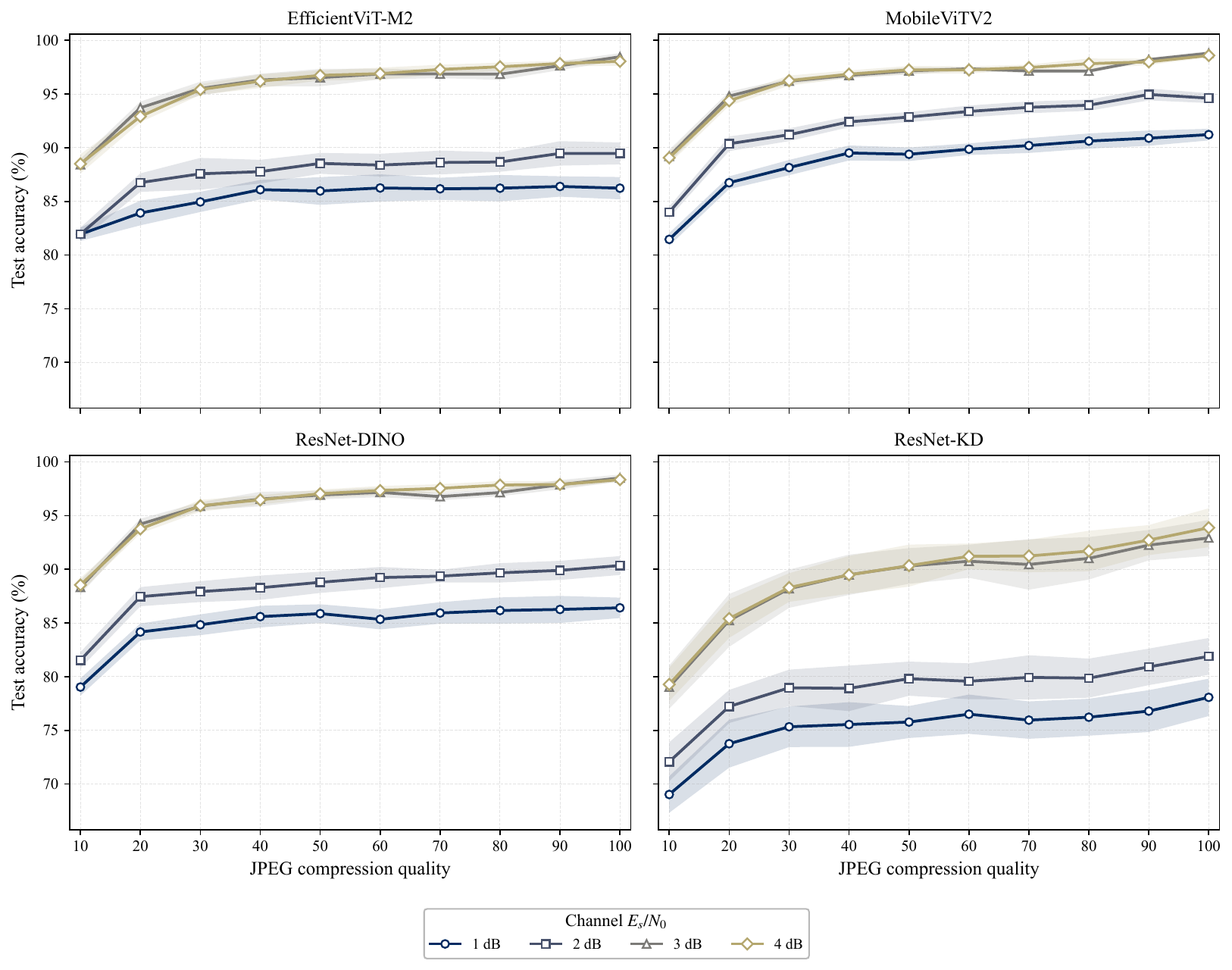}
    \captionsetup{font=small}
    \caption{DVB-S2(X) transmission ablation on EuroSAT: classification accuracy versus JPEG compression quality, one curve per channel SNR ($E_s/N_0$), for four onboard-relevant models. Shaded bands denote $\pm1\sigma$ over five runs.}
    \label{fig:dvbs2x}
    \vspace{-3mm}
\end{figure*}

Fig.~\ref{fig:dvbs2x} reports the DVB-S2(X) transmission ablation, in which four onboard-relevant models (EfficientViT-M2, MobileViTV2, ResNet-DINO, and ResNet-KD, the latter a knowledge-distilled ResNet student from our prior work \cite{le2025semantic, le2026gluse}) are evaluated on EuroSAT across four channel-SNR levels ($E_s/N_0\in\{1,2,3,4\}$~dB) and ten JPEG compression-quality levels, with each operating point averaged over five runs. Accuracy degrades gracefully as channel SNR and compression quality decrease, and EfficientViT-M2 exhibits the smallest degradation span across the grid (a max$-$min accuracy range of $16.53$ points, versus $17.34$ for MobileViTV2, $19.48$ for ResNet-DINO, and $24.90$ for ResNet-KD), remaining the most stable of the four under joint source--channel loss. At the worst corner (1~dB, quality~10) EfficientViT-M2 still attains $81.94\%$, and at the best corner (4~dB, uncompressed) it reaches $98.05\%$. This transmission ablation directly models the input-data uncertainty introduced by the satellite-to-ground link, complementing the sensor-level Gaussian-noise and motion-blur robustness experiments.

\vspace{-2mm}
\subsection{Edge Feasibility and Discussion}

Regarding practical edge feasibility, since flight-representative hardware was not available we assess the onboard cost using platform-independent proxies (parameters, MACs, and model size) rather than device-specific latency; Table~\ref{tab:edge_proxy} summarizes these proxy figures. Relative to a lightweight ResNet-14 baseline ($117.88$~MFLOPs, $10.01$~MB, $93.88\%$ accuracy), EfficientViT-M2 ($203.53$~MFLOPs, $38.19$~MB, $98.76\%$) is $1.73\times$ more compute-intensive and $3.82\times$ larger, but delivers $+4.88$ accuracy points together with the superior noise- and channel-robustness documented above. In absolute terms it remains $\approx$23$\times$ lighter in compute than the CloudScout network already flown onboard $\Phi$-Sat-1, indicating that its footprint is comfortably within a demonstrated onboard budget; direct measurement on flight-representative accelerators (e.g., Intel Myriad VPU, NVIDIA Jetson, or FPGA) is discussed in Section~\ref{sec:limits}.

\begin{table}[!t]
\centering
\footnotesize
\captionsetup{font=small}
\caption{\textsc{Hardware-Agnostic Edge-Feasibility Proxies}}
\label{tab:edge_proxy}

\setlength{\tabcolsep}{4pt}
\begin{tabular}{@{}l r r r@{}}
\toprule
\textbf{Model} & \textbf{MACs} & \textbf{Size (MB)} & \textbf{Acc.\ (\%)} \\
\midrule
ResNet-14                   & 117.88\,M & 10.01  & 93.88 \\
EfficientViT-M2             & 203.53\,M & 38.19  & \textbf{98.76} \\
CloudScout (reconstr.)      & 4.67\,G   & 123.98 & n/a$^{\dagger}$ \\
RepSViT \cite{pang2024repsvit} & 600\,M & --     & n/a$^{\dagger}$ \\
\bottomrule
\multicolumn{4}{@{}p{0.97\columnwidth}@{}}{\scriptsize $^{\dagger}$Different task (cloud detection / onboard image processing), so classification accuracy is not comparable; rows are included for complexity context only. CloudScout figures are from our reconstruction, measured with the same \texttt{fvcore}/\texttt{torchinfo} pipeline as all other models; RepSViT figures are as published \cite{pang2024repsvit}. All reported accuracy values refer to EuroSAT Top-1 classification accuracy.}
\end{tabular}
\vspace{-2mm}
\end{table}

Taken together, the discussion above situates our results against the onboard-EO literature ($\Phi$-Sat-1/CloudScout, $\Phi$-Sat-2, and RepSViT), characterizes the principal input-data uncertainties (native spatial resolution, acquisition noise, and downlink channel/compression loss), and frames the augmentation ablation as a sensitivity analysis over the training-time augmentation policy while optimizer and schedule were held fixed across models. The remaining open issues, and the study's scope boundaries, are consolidated in Section~\ref{sec:limits}.

Lastly, these findings are consistent with the companion end-to-end satellite communications (SatCom) study \cite{nguyen2025semantic, nguyen2026semantic}, in which EfficientViT-M2, evaluated through a DVB-S2(X) downlink emulation with realistic channel coding, RF impairments, and compression losses, sustained above 97\% accuracy under favourable transmission conditions and degraded only gradually as source and channel losses increased, further validating its practicality for bandwidth- and noise-constrained onboard EO mission deployments.

\vspace{-2mm}
\section{Limitations and Future Work} \label{sec:limits}

This section consolidates the aspects of the onboard-EO problem that the present study does not resolve, and states for each of them the research direction it motivates.

The first limitation concerns spectral coverage and the input uncertainties that our degradation models do not represent. The study deliberately uses only the three RGB bands to remain compatible with three-channel ImageNet-pretrained backbones and RGB state-of-the-art baselines. The cost of this choice is visible in our own confusion matrices: the dominant residual errors (River versus Highway; Pasture versus Herbaceous Vegetation) are exactly those that NIR-derived indices such as NDWI and NDVI are designed to resolve. Quantifying how much of this residual error multispectral input recovers requires multispectral-native backbones and is left to future work. Similarly, while our degradation models cover sensor noise, platform-induced blur, and downlink channel/compression loss, they do not span every input uncertainty of operational EO: cloud contamination remains unmodeled, and the effects of varying ground-sampling distance and acquisition gaps are not experimentally isolated. Cloud-aware pre-filtering (e.g., a lightweight CloudScout-style detector cascaded ahead of the classifier) and resolution-robustness studies are natural extensions of this work.

A second limitation concerns hardware realism. All efficiency figures reported here are either platform-independent proxies (parameters, MACs, model size; Table~\ref{tab:edge_proxy}) or measurements on workstation GPUs; no flight-representative silicon was available. Deployment on radiation-tolerant accelerators involves quantization, operator support, and compilation constraints that can change latency and power non-trivially, so our GPU power ranking should be read as indicative rather than device-exact. Benchmarking EfficientViT-M2 on Intel Myriad-class VPUs, NVIDIA Jetson, or space-grade FPGAs is the most direct next step; a complementary route is neuromorphic edge hardware, on which our companion ResNet-GLUSE model runs inference at 852.30~mW on the Akida Brainchip platform \cite{le2026gluse}, although such deployment currently favors architecturally simple CNNs over the attention-based ViTs studied here. Relatedly, the CloudScout complexity figures derive from our reconstruction of the published architecture: although the parameter count was verified layer by layer and all models share one measurement pipeline, the flight build may differ in implementation details, so the $\approx$23$\times$ compute ratio is a faithful approximation rather than an exact flight comparison. For the same reason, a same-task, same-hardware benchmark against the $\Phi$-Sat-1 and $\Phi$-Sat-2 workloads, which alone would justify claims about relative speed or accuracy, remains an open task for future work.

A third limitation concerns the scope of the training and evaluation protocol. To avoid confounding the comparison, optimizer family, learning-rate schedule, and the 25-epoch budget were held fixed across all models; the reported ranking is therefore conditional on this shared configuration. Per-model hyperparameter tuning could shift individual scores, and a full per-model sweep was computationally out of scope, although the margins between the leading compact ViTs and the from-scratch baselines are large enough that a rank reversal appears unlikely; the augmentation ablation should be read as a sensitivity analysis over the data pipeline only. The evaluation is also single-domain and fully supervised: robustness to sensor- and geography-induced domain shift is untested, making unsupervised domain adaptation \cite{binujose2026uda, li2025classaware} a promising complement to our selection methodology. Finally, the study addresses single-task classification only; multitask pre-training such as MTP \cite{wang2024mtp} reaches slightly higher accuracy (99.30\%) but at over 300~M parameters, and extending compact ViTs to multitask settings could balance additional capability against the resource budget available onboard.

Two further directions extend the methodology beyond the setting studied here. Dynamic wireless environments may eventually require on-device adaptation to time-varying transmission errors, which our fixed-model, emulated-channel study does not address. Beyond RGB classification, multimodal DL frameworks \cite{yao2023extended, roy2024cross}, diffusion-based cross-modal representations \cite{tang2024aerogen}, and lightweight state-space sequence models \cite{yao2024spectralmamba} offer routes to richer yet resource-efficient EO processing, and are complementary to the energy-aware model-selection methodology established and validated in this study.

\vspace{-2mm}
\section{Conclusions} \label{sec:concl}

This paper presented an extensive comparison of CNN-based, ResNet-based, and Transformer-based models for RS-IC, evaluating pre-trained ViT architectures for the best balance of complexity, efficiency, and performance, with particular attention to robustness under the noisy inference conditions of satellite EO. The results identify EfficientViT-M2 as the optimal model for onboard deployment: it combines superior robustness to Gaussian noise and motion blur with low computational complexity (203.53~MFLOPs) and a compact size (38.19~MB), offering substantial efficiency gains over models flown on pioneering missions such as $\Phi$-Sat-1 and $\Phi$-Sat-2 while maintaining high classification accuracy with minimal overhead. Under an end-to-end DVB-S2(X) downlink emulation it further shows the most graceful degradation of the models compared, supporting its practicality from sensor capture to ground delivery. The open issues consolidated in Section~\ref{sec:limits}, namely multispectral input, flight-representative hardware, domain shift, and multitask extension, define the roadmap for carrying this energy-aware selection methodology toward operational onboard deployment.

\IEEEtriggeratref{50}
\bibliographystyle{IEEEtran}
\bibliography{IEEEabrv,Bibliography}

\end{document}